\documentclass[10pt,twocolumn,letterpaper]{article}

\usepackage{cvpr}              %

\newif\ifarxiv
\newif\ifnotarxiv
\notarxivfalse
\arxivtrue

\usepackage{graphicx}
\usepackage{graphbox}

\usepackage{amsmath}
\usepackage{amssymb}
\usepackage{appendix}
\usepackage{booktabs}
\usepackage{xcolor}
\usepackage{xspace}
\usepackage{tabularx}
\usepackage{comment}
\usepackage{tablefootnote}
\usepackage{stfloats}
\usepackage{pifont}
\ifnotarxiv
\usepackage[accsupp]{axessibility}  %
\fi

\ifarxiv
\usepackage{axessibility}  %
\fi

\newcommand{\model}{SSCD\xspace}
\newcommand{\modelx}{SSCD}
\newcommand{\uAP}{$\mu AP$\xspace}
\newcommand{\uAPSN}{$\mu AP_{SN}$\xspace}

\DeclareMathOperator{\similar}{sim}

\newcommand{\ed}[1]{{\color{orange}[\textbf{Ed}:#1]}}
\newcommand{\matthijs}[1]{{\color{blue}[\textbf{Matthijs}:#1]}}
\newcommand{\sreya}[1]{{\color{red}[\textbf{Sreya}:#1]}}

\usepackage[pagebackref,breaklinks,colorlinks]{hyperref}

\usepackage[capitalize]{cleveref}
\crefname{section}{Sec.}{Secs.}
\Crefname{section}{Section}{Sections}
\Crefname{table}{Table}{Tables}
\crefname{table}{Tab.}{Tabs.}

\def\cvprPaperID{6654} %
\def\confName{CVPR}
\def\confYear{2022}

\makeatletter
\renewcommand{\paragraph}{%
  \@startsection{paragraph}{4}%
  {\z@}{1.2ex \@plus 1ex \@minus .2ex}{-1em}%
  {\normalfont\normalsize\bfseries}%
}
\makeatother

\begin{document}

\title{A Self-Supervised Descriptor for Image Copy Detection}

\newcommand{\spx}{\hspace{8mm}}

\author{Ed Pizzi \spx Sreya Dutta Roy \spx Sugosh Nagavara Ravindra \spx Priya Goyal \spx Matthijs Douze  \medskip
 \\
{Meta AI}
}
\maketitle

\begin{abstract}

Image copy detection is an important task for content moderation.
We introduce \model, a model that builds on a recent self-supervised contrastive training objective.
We adapt this method to the copy detection task by changing  the architecture and training objective, including a pooling operator from the instance matching literature, and adapting contrastive learning to augmentations that combine images.
Our approach relies on an entropy regularization term, promoting consistent separation between descriptor vectors, and
we demonstrate that this significantly improves copy detection accuracy.
Our method produces a compact descriptor vector, suitable for real-world web scale applications.
Statistical information from a background image distribution can be incorporated into the descriptor.

On the recent DISC2021 benchmark, \model is shown to outperform both baseline copy detection models and self-supervised architectures designed for image classification by huge margins, in all settings.
For example, \model outperforms SimCLR descriptors by 48\% absolute.

Code is available at
\url{https://github.com/facebookresearch/sscd-copy-detection}.

\end{abstract}

\section{Introduction}
\label{sec:intro}

All online photo sharing platforms use content moderation to block or limit the propagation of images that are considered harmuful: terrorist propaganda, misinformation, harassment,
pornography, \emph{etc}. 
Some content moderation can be performed automatically, for unambiguous data like pornographic pictures, but this is much harder for complex data like memes~\cite{kiela2020hateful} or misinformation~\cite{allcott2019trends}.
In these cases, content is moderated manually.
For of viral images, where copies of same image may be uploaded thousands of times, manual moderation of each copy is tedious and unnecessary.
Instead, each image for which a manual moderation decision is taken can be recorded in a database, so that it can be re-identified later and handled automatically.

\begin{figure}
    \hspace*{-1em}
    \includegraphics[width=1.1\columnwidth]{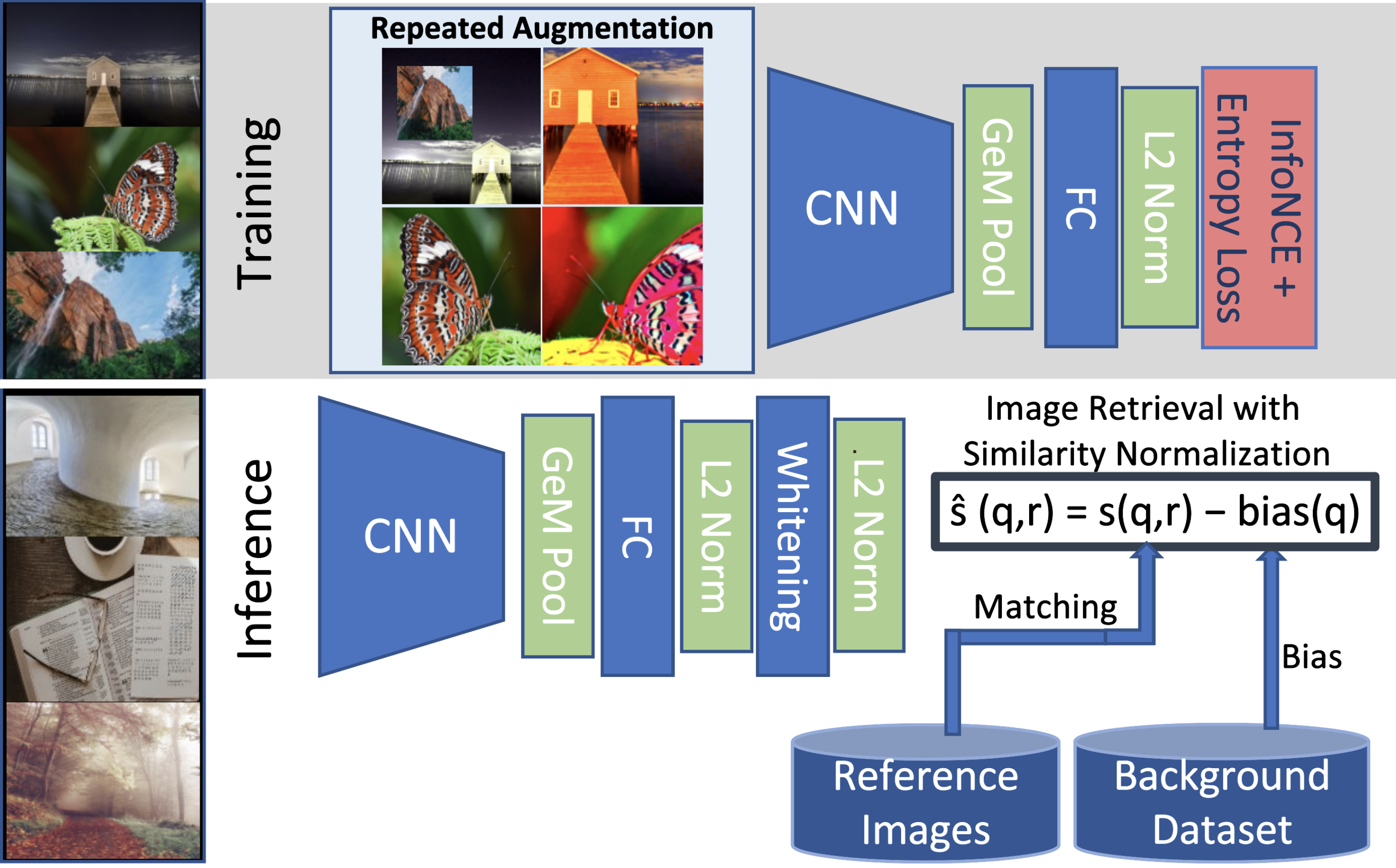}
    \caption{
        The \model architecture for image copy detection. It is based on SimCLR, with the following additions: the entropy regulatization, 
        cutmix/mixup-aware InfoNCE,
        and inference-time score normalization.
    }
    \label{fig:splash}
\end{figure}

This paper is concerned with this basic task of re-identifcation.
This is non trivial because copied images are often altered, for technical reasons (\eg a user shares a mobile phone screenshot that captures additional content), 
or users may make adversarial edits to evade moderation.

Image re-identification is an image matching problem, with two additional challenges. 
The first is the enormous scale at which copy detection systems are deployed.
At this scale, the only feasible approach is to represent images as short descriptor vectors, that can be searched efficiently with approximate nearest neighbor search methods~\cite{johnson2017billion,guo2020accelerating}.
Copy detection systems typically proceed in 2~stages: 
a \emph{retrieval} stage that produces a shortlist of candidate matches and 
a \emph{verification} stage, often based on local descriptor matching that operates on the candidates.
In this work, we are concerned with the first stage.
Figure~\ref{fig:splash} shows the overall architecture of our Self Supervised Copy Detection (\model) approach.

The second challenge is that there is a hard match/non-match decision to take, and positive image pairs are rare.
We wish to limit verification candidates using a threshold, which is a harder constraint than the typical image retrieval setting, where only the order of results matter.

\model uses differential entropy regularization~\cite{sablayrolles2019spreading} to promote a uniform embedding
distribution, which has three effects: 
(1) it makes distances from different embedding regions more comparable;
(2) it avoids the embedding collapse described in~\cite{jing2021collapse}, making full use of the embedding space;
(3) it also improves ranking metrics that do not require consistent thresholds across queries.

Score normalization is important for ranking systems. 
An advanced score normalization relies on matching the query images with a  set of background images. 
In this work, we show how this normalization can be incorporated in the image descriptor itself.
We anticipate that this work will set a strong single-model baseline for image copy detection. We plan to release code and models for our method.

Section~\ref{sec:related} discusses works related to this paper. 
Section~\ref{sec:motivation} motivates the use of an entropy loss term in a simplified setting. 
Section~\ref{sec:method} carefully describes \model.
Section~\ref{sec:experiments} presents results and ablations of our method.
Section~\ref{sec:discussion} points out a few observations about the copy detection task.

\section{Related work}
\label{sec:related}

\paragraph{Content tracing approaches.}

Content tracing on a user-generated photo sharing platform aims at re-identifying images when they circulate out and back into the platform.
There are three broad families of tracing methods: metadata-based~\cite{metadataTest2019,aythora2020multi}, watermarking~\cite{cox2007digital,urvoy2014perceptual,zhu2018hidden,luo2020distortion} and content-based. 
This work belongs to this last class.

%
Classical image datasets for content tracing, like Casia~\cite{dong2013casia,pham2019hybrid} focus on image alterations 
like splicing, removal and copy-move transformations~\cite{team2017nimble,dong2013casia,wen2016coverage} that
alter only a small fraction of the image surface, so the re-identification is done reliably with simple interest-point based techniques.
The challenge is to detect the tampered surface, which is typically approached with deep models inspired by image segmentation~\cite{zhou2018learning,nguyen2021oscar}.
A related line of research is image phylogeny: the objective is to identify the series of edits that were applied to an image between an initial and a final state~\cite{dias2011image,dias2013large,moreira2018image}. 
The Nimble/Media forensics series of competitions organized by NIST aim at benchmarking these tasks~\cite{yates2017nimble,robertson2019manipulation}.
In this work we focus on the identification itself, with strong transformations and near duplicates that need to be distinguished (see Figure~\ref{fig:imageexamples}). 

\paragraph{Semantic and perceptual image comparison}

Several definitions of near-duplicate image matching, form a continuum between pixel-wise copy and instance matching~\cite{jinda2013california,douze2021isc}. 
The definition we use in this work is: images are considered copies iff they come from the same 2D image source. 
More relaxed definitions allow, for example, to match nearby frames in a video. 

There is a large body of literature about solving instance matching~\cite{nister2006scalable,chum2007total,jegou2008hamming,tolias2013aggregate,tolias2016image,radenovic2018revisiting,berman2019multigrain,tolias2020learning} \ie, recognizing images of the same 3D object with viewpoint/camera changes. 
In this work, we build on this literature because it addresses complex image matching,
and
to our knowledge, recent works and benchmarks for strict copy detection are rare~\cite{Douze2009EvaluationOG,wang2015instre}.

\newcommand{\igcomp}[1]{\includegraphics[width=0.34\columnwidth,align=c]{fig/retreieval_examples/compressed/#1.jpg}}
\newcommand{\igcompv}[1]{\includegraphics[height=0.34\columnwidth,align=c]{fig/retreieval_examples/compressed/#1.jpg}}

\newcommand{\imcreds}[1]{\raisebox{0.5\depth}{\scalebox{0.4}{\rotatebox{270}{#1}}}}

\begin{figure}
    \centering
    \hspace*{-1em}
\begin{tabular}{c@{\hspace{1mm}}c@{\hspace{1mm}}c}
    \toprule
    Query  & \model & SimCLR \\
    \midrule
    \smallskip \igcomp{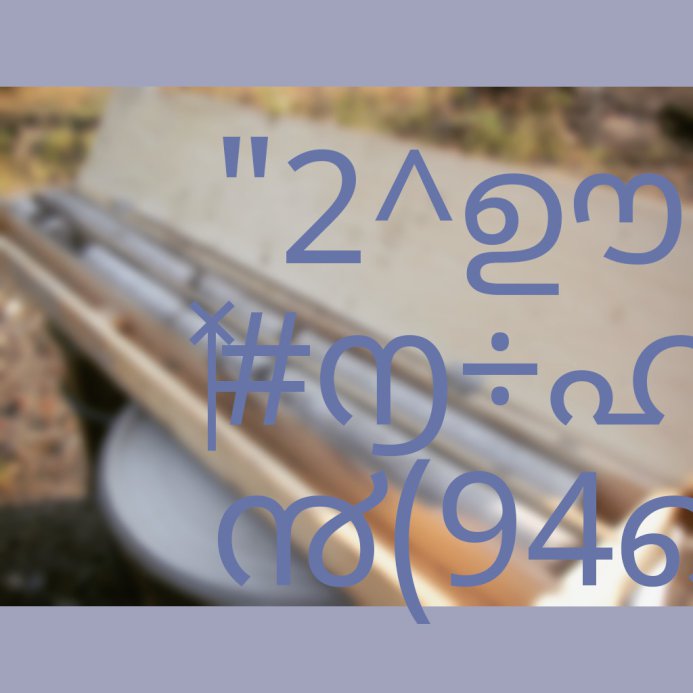} & \igcomp{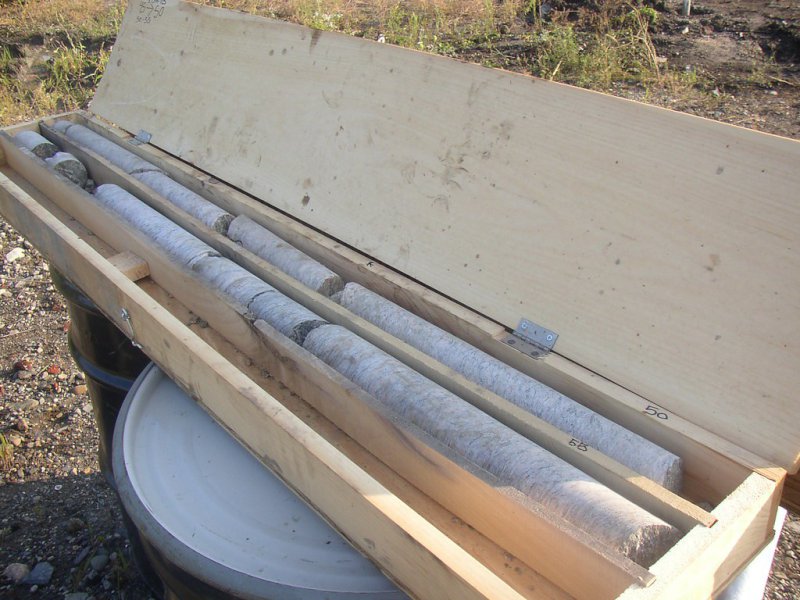}\imcreds{Massachusetts Dept. of Environmental Protection} & \igcompv{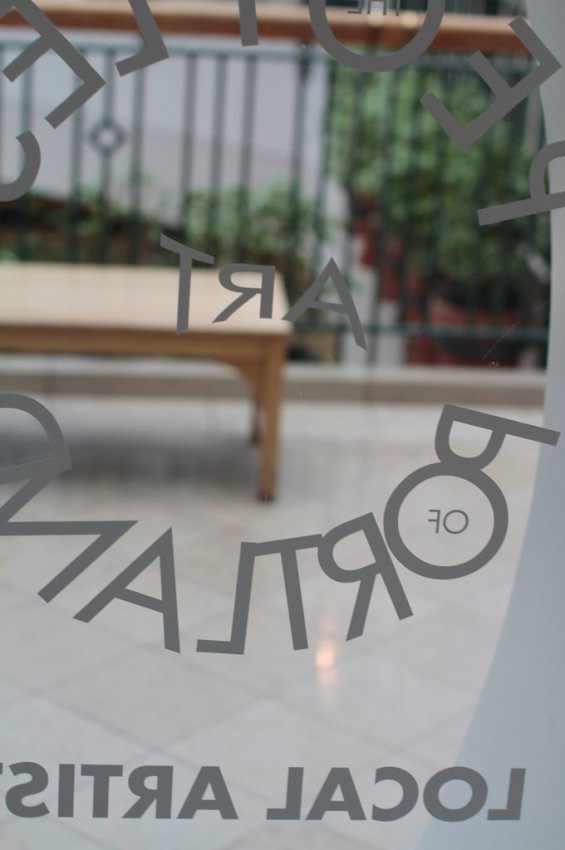}\imcreds{Parker Knight} \\ 
    \smallskip \igcomp{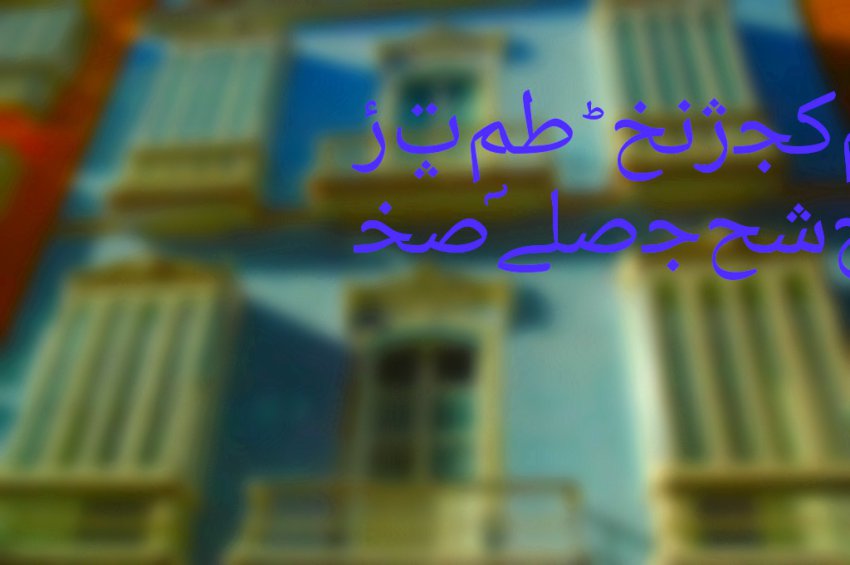} & \igcomp{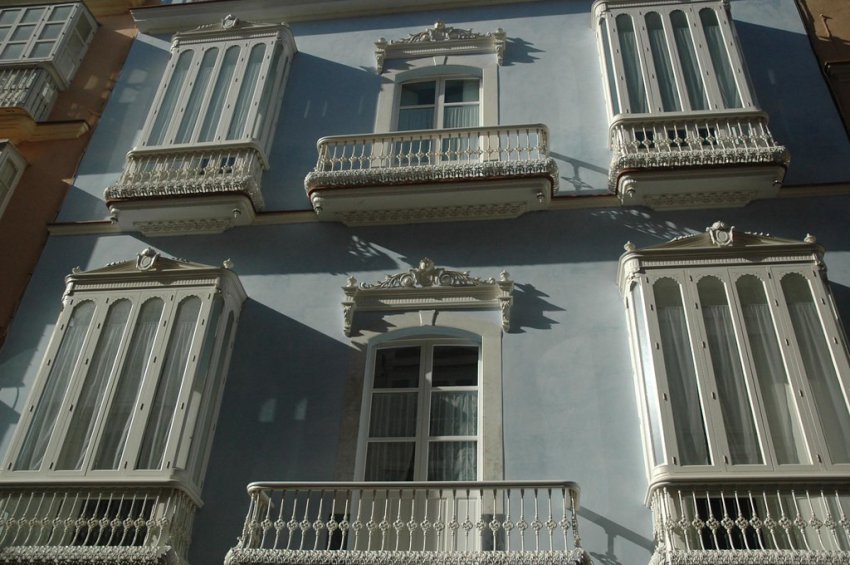}\imcreds{weisserstier} & \igcomp{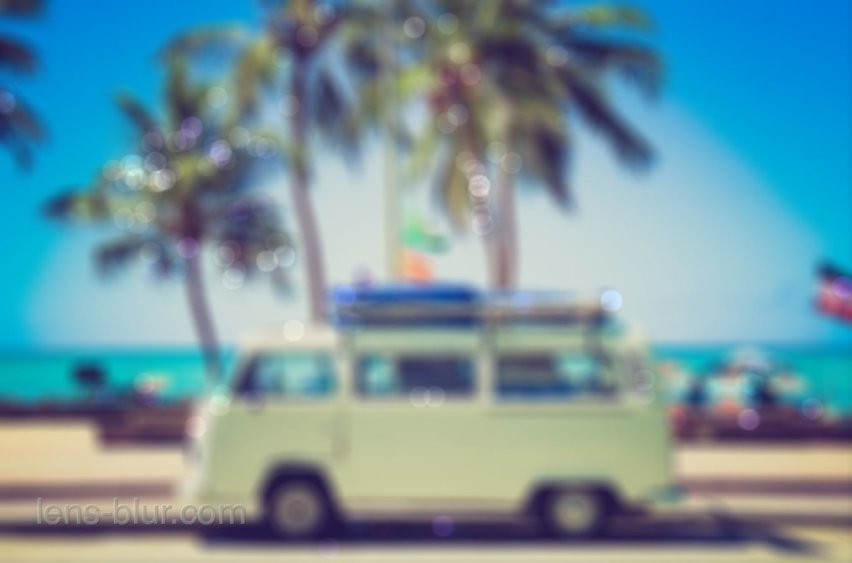}\imcreds{ZHANG HONGFENG} \\
    \smallskip \igcompv{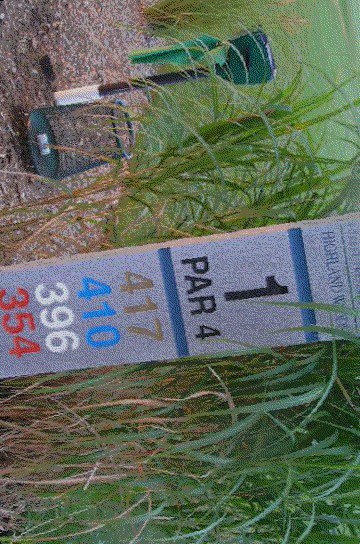} & \igcomp{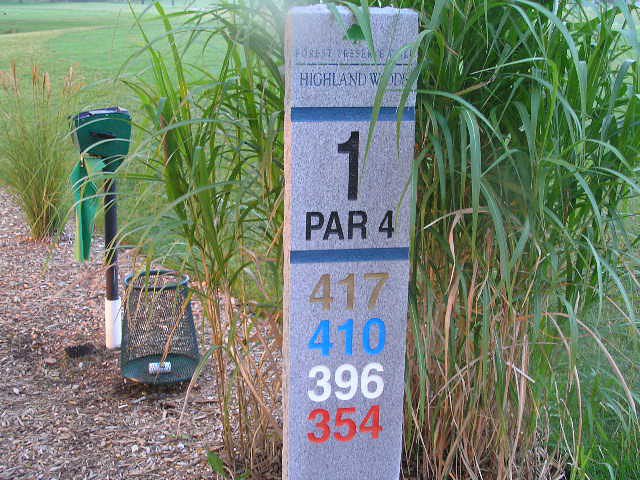}\imcreds{danperry.com} & \igcomp{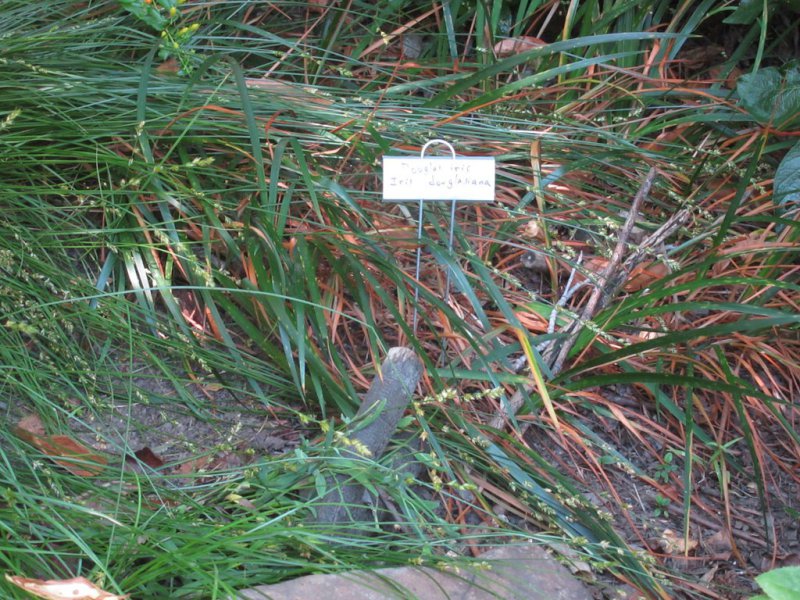}\imcreds{jaycross} \\
    \smallskip \igcomp{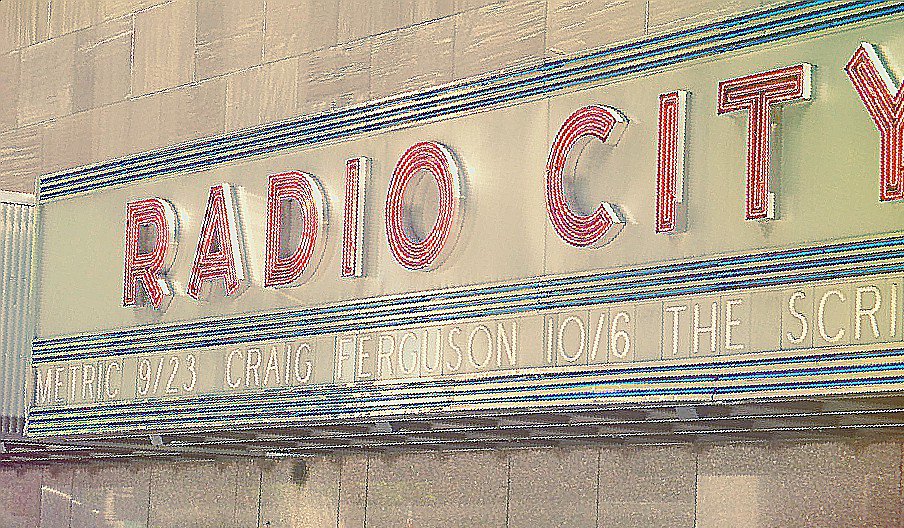} & \igcomp{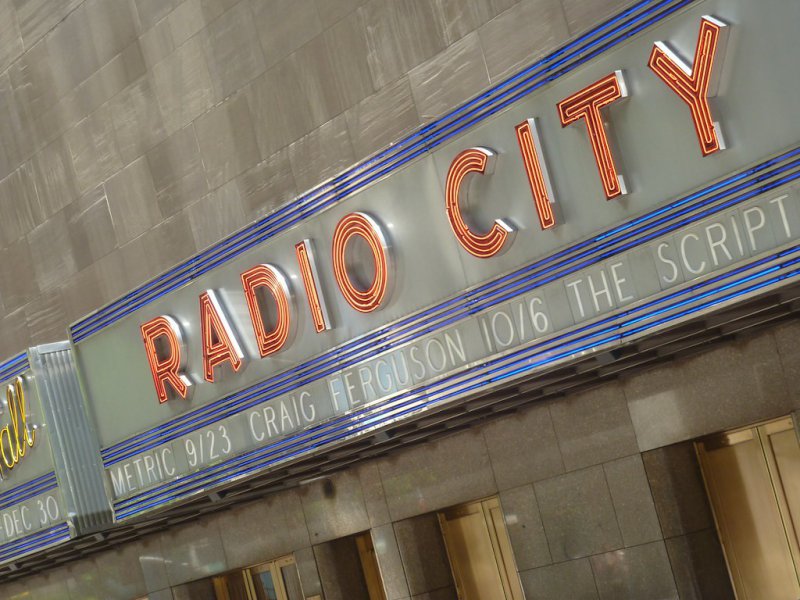}\imcreds{Christopher S Head} & \igcomp{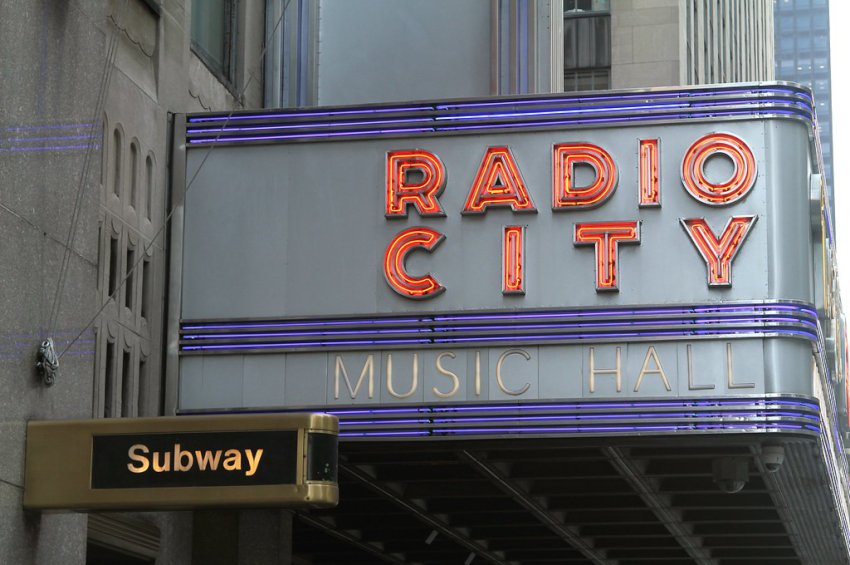}\imcreds{Jim, the Photographer} \\    
    \bottomrule
\end{tabular}
    \caption{
    Example retrieval results from the DISC2021 dataset.
    Each row is an example. From left to right: 
    query image, first result returned by \model, first result returned by the SimCLR baseline. 
    }
    \label{fig:imageexamples}
\end{figure}

\paragraph{Instance matching.}

Classical instance matching relies on 3D matching tools, like interest points~\cite{sivic2003video,nister2006scalable,jegou2008hamming}. 
CNN-based approaches use backbones from image classification, either pre-trained~\cite{babenko2014neural,gong2014multi,tolias2015particular} or trained end-to-end~\cite{radenovic2016bow,Gordo2016DeepIR}, with two adaptations: 
(1) the pooling layer that converts the last CNN activation map to a vector is a max-pooling~\cite{tolias2015particular}, or more generally GeM pooling~\cite{radenovic2018fine}, a form of $L_p$ normalization where $p$ is adapted to the image resolution~\cite{berman2019multigrain};
(2) careful normalization of the vectors. 
In addition to simple L2-normalization~\cite{babenko2014neural},
``whitening'' is often used to compare descriptors~\cite{jegou2012negative,tolias2015particular}. 
An additional normalization technique contrasts the distances \wrt a background distribution of images~\cite{jegou2011exploiting,douze2021isc}. 
In this work, we apply these pooling and normalization techniques to copy detection.

\paragraph{Contrastive self-supervised learning.}

A recent line of self-supervised learning research uses contrastive objectives
that learn image representations that bring transformed images together. %
These methods either discriminate image features~\cite{he2020momentum, chen2020simple, grill2020bootstrap} or the cluster assignments of these image features~\cite{caron2020unsupervised}. 
These methods either rely on memory banks~\cite{he2020momentum, wu2018unsupervised} or large batch sizes~\cite{chen2020simple}.
In particular, SimCLR~\cite{chen2020simple} uses matching transformed image copies as a surrogate task to learn a general image representation that transfer well to other tasks, such as image classification.
A contrastive InfoNCE loss~\cite{oord2018cpc} is used to map copies of the same source image nearby in the embedding space.

\paragraph{Differential entropy regularization.}

Increasing the entropy of media descriptors forces them to spread over the representation space. 
Sablayrolles \etal~\cite{sablayrolles2019spreading} observed that the entropy can be estimated locally with the Kozachenko-Leononenko differential entropy estimator~\cite{beirlant97entropy}, that can be incorporated directly into the loss to maximize descriptor entropy.
The work of El-Nouby \etal~\cite{elnouby2021vitretrieval} is closest to our approach. 
It adds the entropy term to a contrastive loss at fine-tuning time to improve the accuracy for category and instance retrieval.
Our approach is similar, applied to a self-supervised objective and image copy detection.

\section{Motivation}
\label{sec:motivation}

In this section, we start from the SimCLR~\cite{chen2020simple} method, then perform a simple experiment where we combine it with the entropy loss from~\cite{sablayrolles2019spreading} and witness how it impacts classification and copy detection tasks.

\subsection{Preliminaries: SimCLR}
\label{sec:simclrbasics}

SimCLR training is best described at the mini-batch level.
For batches of $N$ images, it creates two augmented copies of each image (repeated augmentations), yielding $2N$ transformed images.
The positive pairs of matching images are $P=\{(i, i + N), (i + N, i)\}_{ i=1..N}$.
We denote positive matches for image $i$ as $P_i = \{j \mid (i, j) \in P\}$.
Each image is transformed by a CNN backbone network. %
The final activation map of the CNN is average pooled, then projected using a two-layer MLP into a L2-normalized descriptor
$z_i \in \mathbb{R}^d$.
Descriptors are compared with a cosine similarity: $\similar(z_i, z_j) = z_i^\top z_j$.
A contrastive InfoNCE loss maximizes the similarity between copies relative to
the similarity of non-copies.
For inference (\eg to transfer to image classification), SimCLR discards the training-time MLP, using globally pooled features from the CNN trunk directly.

\paragraph{The InfoNCE loss.}

SimCLR's InfoNCE loss is a softmax cross-entropy with temperature, that  matches descriptors to other descriptors. 
Let $s_{i,j}$ be the temperature-adjusted cosine similarity $s_{i,j}=\similar(z_i, z_j) / \tau$.
The InfoNCE loss is defined as a mean of $\ell_{i,j}$ terms for positive pairs $(i, j) \in P$:
\begin{equation}
    \ell_{i,j} = - \log \frac{\exp( s_{i,j} )}{ \sum_{k \ne i} \exp( s_{i,k} ) } 
    \label{eq:infonce_term}
\end{equation}
\begin{equation}
    \mathcal{L}_\mathrm{InfoNCE} = \frac{1}{|P|} \sum_{i,j \in P} \ell_{i,j}.
\label{eq:infonceloss}
\end{equation}

\subsection{Entropy regularization}

We use the differential entropy loss proposed in \cite{sablayrolles2019spreading}, based on the Kozachenko-Leononenko estimator.
We adapt it to the repeated augmentation setting by only regularizing neighbors from different source images:
\begin{equation}
\mathcal{L}_\mathrm{KoLeo} = - \frac{1}{N} \sum_{i=1}^N{\log\big(\min_{j \not\in \hat{P}_i} \left\lVert z_i - z_j \right\rVert \big)},
 \label{eq:basicentropy}
\end{equation}
where $\hat{P}_i = P_i \cup \{i\}$. %
Since this entropy loss is a log of the distance to the nearest neighbor, its impact is very high for nearby vectors but dampens quickly when the descriptors are far apart.
The effect is to ``push'' apart nearby vectors. 

\begin{figure}[h]
    \centering
    \includegraphics[width=0.95\columnwidth]{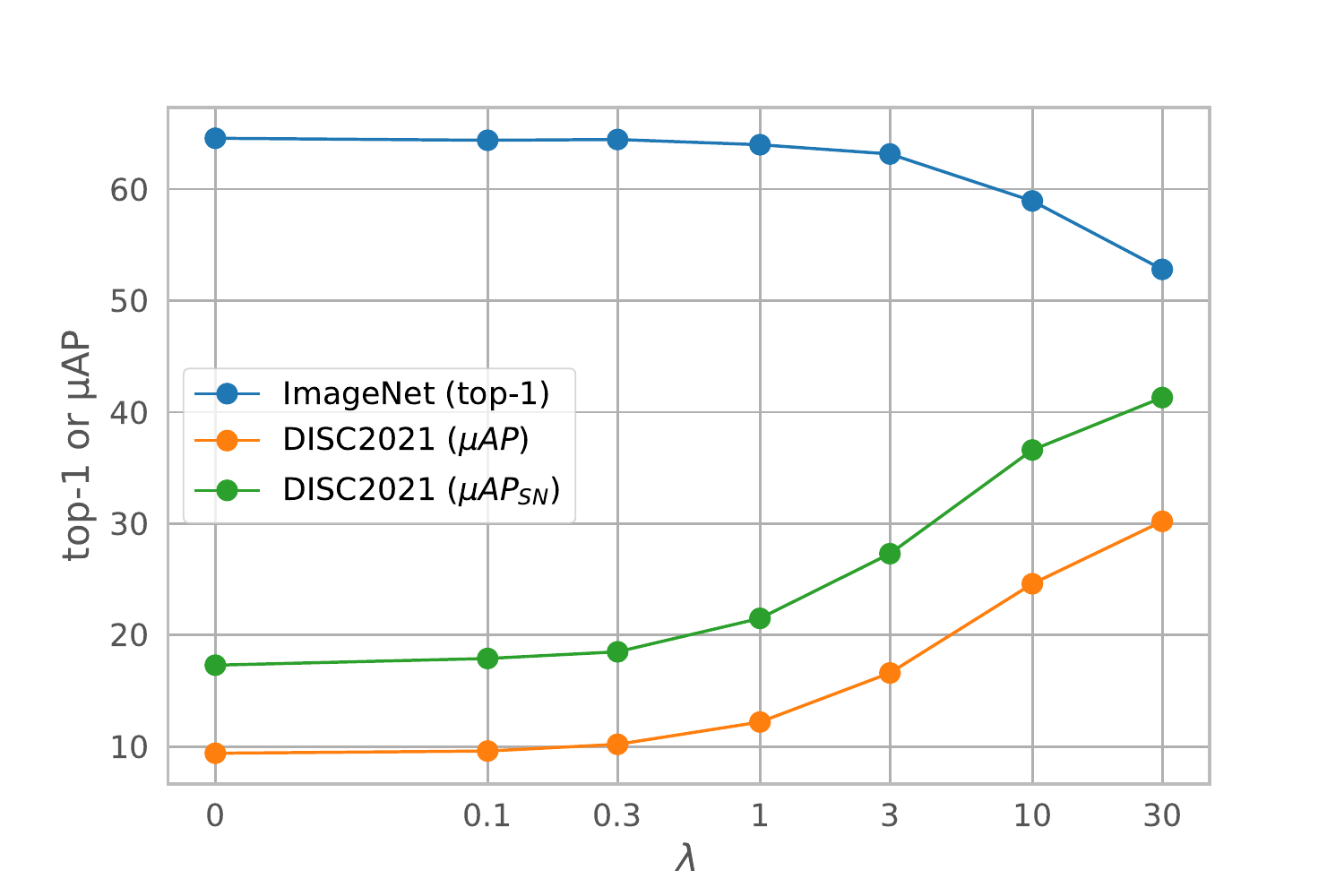}
    \caption{
    Preliminary experiment:
    we train SimCLR models on ImageNet with varying
    differential entropy regularization strength,  
    (regular SimCLR: $\lambda=0$).
    We measure: 
    ImageNet linear classification accuracy and 
    DISC2021 micro average precision (\uAP), 
    with optional score normalization (\uAPSN). 
    The ImageNet and DISC2021 measures are not comparable, but trends within each curve are significant.
    }
    \label{fig:simclr_entropy_ablation}
\end{figure}

\subsection{Experiment: SimCLR and entropy}

For this experiment, we combine our contrastive loss with the entropy loss, using a weighting factor $\lambda$, similar to~\cite{sablayrolles2019spreading,elnouby2021vitretrieval}: 
\begin{equation}
    \mathcal{L}_\mathrm{basic} = \mathcal{L}_\mathrm{InfoNCE} + \lambda \mathcal{L}_\mathrm{KoLeo}.
    \label{eq:combinedloss_prelim}
\end{equation}

We then evaluate the impact of the combined loss on an image classifcation setting and a copy detection setting, see Section~\ref{sec:datasets} for more details about the setup.

Figure~\ref{fig:simclr_entropy_ablation} shows how
varying entropy loss weight $\lambda$ impacts both tasks.
As the entropy loss weight increases, ImageNet linear classification accuracy decreases: this loss term is not helpful for classification. 
Conversely, for copy detection the accuracy increases significantly.  

\begin{figure}
    \centering
    \includegraphics[width=\columnwidth]{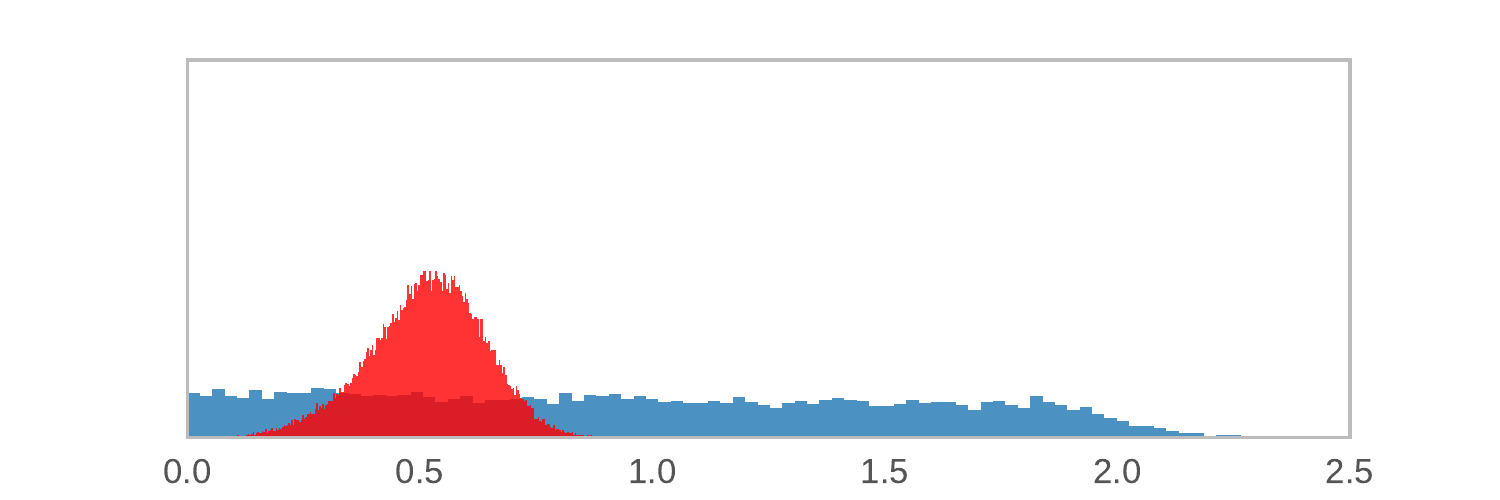}
    \includegraphics[width=\columnwidth]{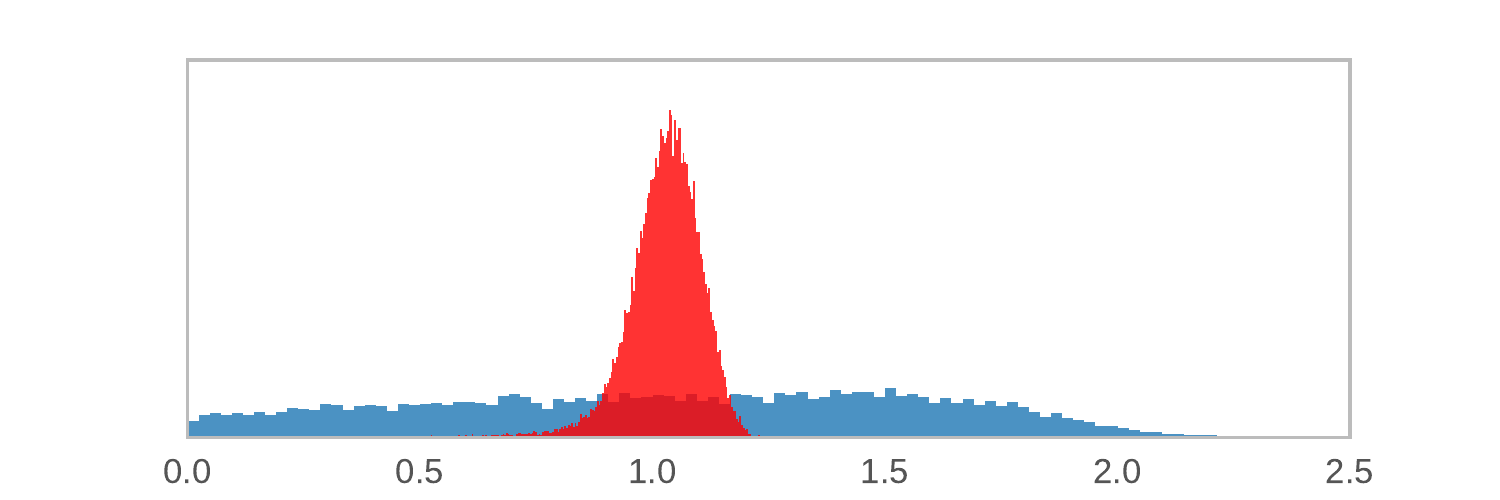}
    \caption{
    Preliminary experiment: histogram of squared distances for DISC2021 {\color{blue}matching images} and {\color{red}non-matching nearest neighbors}.
    Above: baseline SimCLR. Below: SimCLR combined with entropy regularization (weight $\lambda=30$), without whitening or similarity normalization.
    }
    \label{fig:histograms}
\end{figure}

Figure~\ref{fig:histograms} shows the distribution of distances between matching images (positive pairs) and the nearest non-matching neighbors (negative pairs). %
Applying the entropy loss increases all distances and makes the negative distance distribution more narrow. 
The result is that there is a larger contrast between positive pairs and the mode of the negative distribution, \ie they are more clearly separated.

\section{Method}
\label{sec:method}

Having seen how the entropy loss improves copy detection accuracy,
in this section we expand it into a robust image copy detection approach: \model. 
This entails adapting the architecture, the data augmentation, the pooling and adding a normalization stage, as shown in Figure~\ref{fig:splash}.

%

%

\subsection{Architecture}

\model uses a ResNet-50 convolutional trunk to extract image features. 
We standardize on this architecture because it is widely used, well optimized and still very competitive for image classification~\cite{wightman2021resnet}, but any CNN or transformer backbone could be used (see Section~\ref{sec:experiments}).

\paragraph{Pooling.}

For classification, the last CNN activation map is converted to a vector by mean pooling.
We use generalized mean (GeM) pooling instead, which was shown~\cite{radenovic2018fine,berman2019multigrain} to improve the discriminative ability of descriptors.
This is desirable for instance retrieval and our copy detection case alike. 
GeM introduces a parameter $p$, equivalent to average pooling when $p=1$ and max-pooling when $p\rightarrow \infty$.
\model uses $p = 3$, following common practice for image retrieval models~\cite{tolias2015particular,radenovic2018fine,berman2019multigrain}.

While GeM pooling at inference time systematically improves accuracy, 
we observe that it is beneficial at training time only in combination with the differential entropy regularization, \ie
with a vanilla InfoNCE it is better to train with average pooling.
We conjecture that GeM pooling may reduce the difficulty of the training task
without the additional objective of maximally separating embedding points.
We observe that learning the scalar $p$, as proposed in~\cite{radenovic2018fine},
fails for contrastive learning: the pooling parameter grows unbounded %
until training becomes numerically unstable.

\paragraph{Descriptor projection.}

SimCLR uses a 2-layer MLP projection at training time.
For inference, the MLP is discarded and CNN trunk features are used directly.
The MLP is partly motivated to retain transformation-covariant features
in the base network, which may be useful for downstream tasks, despite a training task that requires a transformation-invariant descriptor.
Jing \etal~\cite{jing2021collapse} also
find that the MLP insulates the trunk model from an embedding
collapse into a lower-dimensional space caused by the InfoNCE loss.

For \model, the training and inference tasks are the same, 
obviating the need for
transformation-covariant features, and differential entropy regularization
prevents the dimensional collapse.
We replace the MLP with a simple linear projection to the target descriptor size, and retain this projection for inference.

\subsection{Data Augmentation}

Self-supervised contrastive objectives learn to match images across image transforms.
These methods are sensitive to the augmentations seen at training time~\cite{chen2020simple}, since invariance to these transforms is the only supervisory signal.

\begin{table}
\hspace*{-0.5em}
\scalebox{0.85}{
    \begin{tabular}{|l|l|}
    \hline
    type &  details \\
    \hline
    SimCLR & horizontal flip, random crop, color jitter, grayscale, \\
    & Gaussian blur \\
    Strong blur & 50\% large-radius Gaussian blur ($\sigma \in [1, 5]$)\\
    Advanced & 10\% rotation, 10\% text, 20\% emoji, \\
    & 20\% JPEG compression \\
    Adv. + mixup & 2.5\% mixup, 2.5\% cutmix \\
    \hline
    \end{tabular}}
    \caption{List of data augmentations used for \model. 
    The presentations is incremental: each set of augmentations includes the ones from all rows before.
    The percentages are probabilities to apply each augmentation.
    }
    \label{tab:augmentations}
\end{table}

Table~\ref{tab:augmentations} lists the \model augmentations used in our experiments. %
Note that since our main evaluation dataset (DISC2021) is built in part with data augmentation, there is a risk of overfitting to the augmentations of that dataset.
This is mitigated by (1) DISC2021's set of augmentations is not known precisely and 
(2) we present strong results trained using a simple blur augmentation.
Our starting baseline is the default set of SimCLR augmentations.

\paragraph{Strong blur.}
Empirically, copy detection benefits from a stronger blur than is typically used
for contrastive learning.
We  strengthen the blur augmentation compared to SimCLR.
We suggest that invariance to blur confers a low-frequency bias, reducing the model's sensitivity to high-frequency noise common to real world copies. %
We use this setting for most ablation steps, because it is easy to
reproduce, and provides a good baseline setting for comparing methods.
This augmentation was initially tuned on a proprietary dataset,
and is unlikely to overfit to DISC2021. 

\paragraph{Advanced augmentations.}

We evaluate our method with additional augmentations, to demonstrate how
\model extends as augmentations are added. 
Half of rotations rotate by multiples of 90 degrees and half are unconstrained. 
The text has a random font, text, opacity, font size, and color. 
We add emoji of random size. 
We apply JPEG compression with randomly sampled compression quality.
These augmentations are somewhat inspired by DISC2021 but are still fairly generic for image copy detection problems.

\paragraph{Mixed images.}

\label{sec:mixup}
We use two augmentations that combine content from two images within a training batch.
In a copy detection context, these augmentations model partial copies, where part of an image is included in a composite image. 
Mixup~\cite{zhang2018mixup} is a pixelwise weighted average of two images ($a$ and $b$) with parameter $\gamma \in [0, 1]$:
$ \gamma \cdot a + (1 - \gamma) \cdot b$.
CutMix~\cite{Yun2019CutMix} moves rectangular regions from one image into another.
See Appendix \ref{sec:implementation_details} for implementation details.
Mixed images match multiple images in the batch, requiring changes to our losses, outlined below.

\subsection{Loss Functions}

\model uses a weighted combination of the contrastive InfoNCE and the entropy loss, as in Equation~(\ref{eq:combinedloss_prelim}).
However, we need to adapt both losses for the mixed-image augmentation case, where $P_i$ may contain multiple matching images.

\paragraph{InfoNCE with MixUp/CutMix augmentations.}

We adapt the InfoNCE loss (see Section~\ref{sec:simclrbasics}) to accommodate augmentations that mix features from multiple images.
Given an image $i$ with full or partial matches $j \in P_i$, we modify the pairwise loss term from Equation~(\ref{eq:infonce_term}) as:
\begin{equation}
    \hat{\ell}_{i,j} = - \log \frac{\exp(s_{i,j})}{
        \exp(s_{i,j} ) +
        \sum_{k \not\in \hat{P}_i} \exp(s_{i,j})
    },
\end{equation}
where $\hat{P}_i = P_i \cup \{i\}$. We then combine these terms by taking a mean per image, so that each image contributes similarly to the overall loss, and average per-image losses. Note that this is equivalent to InfoNCE for non-mixed images.
\begin{equation}
    \mathcal{L}_\mathrm{InfoNCE-mix} = \frac{1}{2N} \sum_{i=1}^{2N} \frac{1}{|P_i|} \sum_{j \in P_i} \hat{\ell}_{i,j}.
\end{equation}

%
%

%

%
%
%
%
%
%
%

%

\paragraph{Entropy loss.}

Our formulation of the entropy loss in Equation~(\ref{eq:basicentropy}) remains the same, with $\hat{P}_i$ updated to include multiple matching images.

\paragraph{Combination.}

%
The losses are combined with entropy weight parameter $\lambda$:
\begin{equation}
    \mathcal{L} = \mathcal{L}_\mathrm{InfoNCE-mix} + \lambda \mathcal{L}_\mathrm{KoLeo}
    \label{eq:combinedloss}
\end{equation}

\paragraph{Multi-GPU implementation.}

The contrastive matching task benefits from a large batch size, since this provides stronger negatives.
Losses are evaluated over the global batch, after aggregating image descriptors across GPUs.
Descriptors from all GPUs are included in the negatives InfoNCE matches against, and we choose nearest neighbors for entropy regularization from the global batch.
Batch normalization statistics are synchronized across GPUs to avoid leaking information within a batch.
We use the LARS~\cite{you2017largebatch} optimizer for stable training at large batch size.

\subsection{Inference and retrieval}

For inference, the loss terms are discarded. 
Features are extracted from the images using the convolutional trunk followed by GeM pooling, the linear projection head, and L2 normalization.
Then we apply whitening to the descriptors.
The whitening matrix 
is learned on the DISC2021 training set.
The descriptors are compared with cosine similarity %
or equivalently with simple L2 distance.

\subsection{Similarity normalization}

We follow \cite{douze2021isc} using similarity
normalization~\cite{jegou2011exploiting,conneau2017word} as one of our evaluation settings.
It uses a background dataset of images as a noise distribution, and produces high similarity scores only for queries whose reference similarity is
greater than their similarity to nearest neighbors in the background dataset.
Given a query image $q$ and a reference image $r$ with similarity $s(q, r) = \similar(z_q, z_r)$,
the adjusted similarity is $\hat{s}_0(q, r) = s(q, r) - \beta s(q, b_n)$ where $b_n$ is the $n$\textsuperscript{th} nearest neighbor from
the background dataset, and $\beta \ge 0$ is a weight. 

\newcommand{\nend}{n_\mathrm{end}}

We generalize this by aggregating an average similarity
across multiple neighbors ($n$ to $\nend$) from the background dataset:
\begin{equation}
    \hat{s}(q, r) = s(q, r) -
    \underbrace{
    \frac{\beta}{1 + \nend - n} \sum_{i = n}^{\nend}{ s(q, b_i) 
    }
    }_{\mathrm{bias}(q)}.
\end{equation}

\paragraph{Integrated bias.}

Carrying around a bias term makes indexing of descriptors more complex. %
Therefore, we include the bias into the descriptors as an additional dimension: 
\begin{equation}
    \hat{z}_q = [ z_q \,\,\, -\mathrm{bias}(q)] 
    \hspace{1cm}
    \hat{z}_r = [ z_r \,\,\, 1] 
\end{equation}
Then we are back to $\hat{s}(q, r)=\similar(\hat{z}_q,\hat{z}_r)$. 
The descriptors are not normalized, \ie the dot product similarity is not equivalent to L2 distance. 
If L2 distance is preferred for indexing, it is possible to convert the max dot product search task into L2 search using the approach from~\cite{bachrach2014speeding}.

Similarity normalization consistently improves metrics. 
However it adds operational complexity, and may make it difficult to detect content similar to the background distribution. 
Therefore, we report results both with and without this normalization.

\section{Experiments}
\label{sec:experiments}

In this section we evaluate \model for image copy detection. 
Despite its relative simplicity, it depends on various settings that we evaluate in an extensive ablation study.

\newcommand{\modellarge}{\modelx\textsubscript{large}\xspace}
\newcommand{\SimCLRCD}{SimCLR\textsubscript{CD}\xspace}

\subsection{Datasets}
\label{sec:datasets}

\paragraph{DISC2021.}

Most evaluations are on the validation dataset of the
Image Similarity Challenge, DISC2021~\cite{douze2021isc}.
DISC2021 contains both automated image transforms and manual edits. 
There are 1 million reference images and 50,000 query images, of which 10,000 are
true copies. A disjoint 1 million image training set is used for model training and as background dataset for score normalization. The training
set contains no copies or labels, but is representative of the image distribution
of the dataset.
The performance is evaluated with micro average precision (\uAP) that measures the precision-recall tradeoff with a uniform distance threshold.

\paragraph{ImageNet.}

For some experiments we train models on the ImageNet~\cite{Russakovsky2015ImageNet12} training set (ignoring the class labels). 
We use ImageNet linear classification to measure how our copy detection methods affect semantic representation learning.

\paragraph{Copydays}

\cite{Douze2009EvaluationOG} is a small copy detection dataset.
Following common practice~\cite{berman2019multigrain,caron2021emerging}, we augment it with 10k distractors from YFCC100M~\cite{Thomee2016YFCC100MTN}, a setting known as CD10K, and evaluate the retrieval performance with mean average precision ($mAP$) on the ``strong'' subset of robustly transformed copies.
In addition to this standard measure, we evaluate the \uAP on the overall dataset.

\subsection{Training implementation}

We use the training schedule and hyperparameters from SimCLR~\cite{chen2020simple}:
batch size $N=4096$, resolution $224 \times 224$, learning rate of $0.3 \times N/256$, and a weight decay of $10^{-6}$.
We train models for 100 epochs on either ImageNet or the DISC training
set, using a a cosine learning rate schedule without restarts and with a linear
ramp-up. We use the LARS optimizer for stable training at large batch size.
We train at spatial resolution $224 \times 224$.

We use a lower temperature than SimCLR, $\tau = 0.05$ versus $0.1$,
following an observation in \cite{chen2020simple}
that this setting yields better accuracy on the training task, while reducing
accuracy of downstream classification tasks.

\subsection{Evaluation protocol}

\paragraph{Inference.}
We resize the small edge of an 
image to size 288 preserving aspect ratio for fully convolutional models.
We use a larger inference size than seen at training to avoid train-test discrepancy~\cite{touvron2019fixing}.
We use different preprocessing for the DINO~\cite{caron2021emerging} ViT baseline, following their copy detection method. See Appendix~\ref{sec:implementation_details} for details.

\paragraph{Descriptor postprocessing.}

Image retrieval benefits from PCA whitening.
\model descriptors are whitened, then L2 normalized.
For baseline methods that use CNN trunk features, %
we L2 normalize both before and after whitening.
SimCLR projection features often occupy a low-dimensional subspace, making whitening at full descriptor size unstable, and many representations perform better when whitened with low-variance dimensions excluded.
For baseline methods, we try dimensionalities $\{d, \frac{3}{4} d, \frac{d}{2}, \frac{d}{4}, \ldots \}$ and choose the one that maximizes the final accuracy.
For \model, we whiten at full descriptor size.

We use the FAISS~\cite{johnson2017billion} library to apply embedding postprocessing
and perform exhaustive k-nearest neighbor search. We train PCA on the DISC2021
training dataset, following standard protocol for this dataset.

\subsection{Results}

\renewcommand{\ast}{^*}

\begin{table}[h]
\hspace*{-1.5em}
\resizebox{1.1\columnwidth}{!}{

\begin{tabular}{lllrrr}
\toprule
                                                  method &     trained on &    transforms & dims &   \uAP &  \uAPSN \\
\midrule
 Multigrain~\cite{berman2019multigrain,douze2021isc} &  ImageNet$\ast$ &               & 1500 &  16.5 &    36.5 \\
          HOW~\cite{tolias2020learning,douze2021isc} &  SfM-120k$\ast$ &               &      &  17.3 &    37.2 \\
\midrule
              Multigrain~\cite{berman2019multigrain} &  ImageNet$\ast$ &               & 2048 &  20.5 &    41.7 \\
                       DINO~\cite{caron2021emerging} $^\dagger$ &    ImageNet &               & 1500 &  32.2 &    53.8 \\
                  SimCLR~\cite{chen2020simple} trunk &    ImageNet &        SimCLR & 2048 &  13.1 &    33.9 \\
                   SimCLR~\cite{chen2020simple} proj &    ImageNet &        SimCLR &  128 &   9.4 &    17.3 \\
\midrule
                      \SimCLRCD trunk                &    ImageNet &   strong blur & 2048 &  39.8 &    56.8 \\
                                              \model &    ImageNet &   strong blur &  512 &  50.4 &    64.5 \\
                                              \model &    ImageNet &      advanced &  512 &  55.5 &    71.0 \\
                                              \model &    ImageNet &  adv.+mixup   &  512 &  56.8 &    72.2 \\
\midrule
                                              \model &        DISC &   strong blur &  512 &  54.8 &    63.6 \\
                                              \model &        DISC &      advanced &  512 &  60.4 &    71.1 \\
                                              \model &        DISC &  adv.+mixup   &  512 &  61.5 &    72.5 \\
\midrule
                                              \modellarge$^\dagger$ &        DISC &  adv.+mixup   &  1024 & {\bf 63.7} & {\bf 75.3} \\
\bottomrule
\end{tabular}

}
\caption{
    Copy detection performance in \uAP on the DISC2021 dataset.
    $\ast$: methods that use supervised labels. $^\dagger$: trunk larger than ResNet50.
    DINO baseline uses ViT-B/16.
    \label{tab:results}
}
\end{table}

\paragraph{DISC results.}

Table~\ref{tab:results} reports DISC2021 results from the baseline methods
published in \cite{douze2021isc} and \model.
Our evaluation protocol obtains somewhat stronger results for the Multigrain baseline (3\textsuperscript{rd} row).
The first observation is that \model improves the baseline accuracy by 2$\times$ to 5$\times$ before score normalization, demonstrating
that copy detection benefits from specific architectural and training adaptations. %

We present results on a few different \model models trained on ImageNet 
or DISC2021, using the three
augmentation settings we propose. 
The intermediate model \SimCLRCD has all of our proposed changes
except the entropy loss.
\modellarge model uses a larger descriptor size and a ResNeXt-101 trunk.

We evaluate SimCLR using both trunk and projected features, and find trunk features (\uAP$=13.1$) to
outperform features from the projection head (\uAP$=9.4$) with and without score normalization.
Further experiments (Appendix~\ref{sec:additional_ablations}) show the reverse when training with entropy loss: 
projected features have similar accuracy to trunk features, despite a much more compact representation.

The gain of \SimCLRCD (\uAP$=39.8$ without score normalization) over SimCLR (13.1) is decomposed in Section~\ref{sec:ablations}. 
Introducing the entropy loss in \model contributes an additional 10\% absolute of \uAP, which is further increased by stronger augmentations (+6.2\%) and training on a dataset with less domain shift (+4.7\%).
These findings are confirmed after score normalization.

\paragraph{Copydays results.}

Table~\ref{tab:copydays} %
reports results for baseline methods using publicly released models, but omit Multigrain settings that we were unable to reproduce.
We used published preprocesing settings for baselines and whitening.
Our DINO results outperform published results.

\begin{table}[h]
\centering
\resizebox{\columnwidth}{!}{
\begin{tabular}{llrlrr}
\toprule
      model &     trunk & dims &   size &  $mAP$  & \uAP \\
\midrule
 Multigrain~\cite{berman2019multigrain} &  ResNet50 &       1500 &  long 800 &  82.3 &  77.3 \\
       DINO~\cite{caron2021emerging} &  ViT-B/16 &       1536 &           $224^2$ &  82.8 &  92.3 \\
       DINO~\cite{caron2021emerging} &   ViT-B/8 &       1536 &           $320^2$ &  86.1 &  88.4 \\
\midrule
       \model &  ResNet50 &        512 &  short 288 &  86.6 &  \bf{98.1} \\
\midrule
       \modellarge &  ResNeXt101 &     1024 &  long 800 &  \bf{93.6} &  97.1 \\
\bottomrule
\end{tabular}
} %

    \caption{
    Copydays (CD10K) accuracy measured in mAP on the ``strong'' subset, and \uAP on the full dataset. 
    \label{tab:copydays}
    }
\end{table}

The first \model result is with all settings from our DISC2021 experiments, where we resize the short side of each image to 288 pixels.
With no tuning on this dataset, our method outperforms published results.
We also show results for \modellarge using a ResNeXt101 trunk and 1024 descriptor dimensions, at larger inference size.
We report more results on CD10K in Appendix~\ref{sec:copydays_full}.

In addition to state-of-the-art accuracy using the customary $mAP$ ranking metric, our method provides a significant improvement in the global \uAP metric, indicating better distance calibration.
On high-resolution images that are common for image retrieval, we observe improved $mAP$ but degraded \uAP.
\model descriptors are more compact than baselines.

\subsection{Ablations}
\label{sec:ablations}

\paragraph{Comparison with SimCLR.}

We provide a stepwise comparison between SimCLR and our method in
Table~\ref{tab:imagenet_ablations}. %
SimCLR projection features are not particularly strong for this task until we apply several of our adaptations.
SimCLR is unable to exploit a $\mathbb{R}^{512}$ descriptor, only slightly outperforming its $\mathbb{R}^{128}$ setting.
\SimCLRCD represents our architectural and hyper-parameter changes before adding differential entropy representation.
Differential entropy regularization alone adds +17.4\% \uAP and +12.9\% \uAPSN, more than any other step.

\begin{table}[h]
    \hspace*{-1.5em}
\resizebox{1.1\columnwidth}{!}{

\begin{tabular}{llrrr|rr}
\toprule

& \multicolumn{2}{l}{Score normalization:} & \multicolumn{2}{c}{No} & \multicolumn{2}{c}{Yes} \\
      name &                     method & dims &    \uAP &  256d &  \uAPSN & 256d \\
\midrule
    SimCLR &             trunk features &       2048 &  13.1 &   7.3 &  33.9 &   26.8 \\
           &            + GeM pooling   &       2048 &  21.5 &  12.1 &  45.3 &   35.8 \\
\midrule
    SimCLR &                 projection &        128 &   9.4 &   9.4 &  17.3 &   17.3 \\
           &            + GeM pooling   &        128 &  11.1 &  11.1 &  18.8 &   18.8 \\
           &              + strong blur &        128 &  14.1 &  14.1 &  26.0 &   26.0 \\
           &          + low temp        &        128 &  26.0 &  26.0 &  41.5 &   41.5 \\
           &          + 512d proj       &        512 &  27.5 &  27.5 &  43.5 &   43.5 \\
 \SimCLRCD &        + linear proj   &        512 &  33.0 &  32.4 &  51.6 &   50.5 \\
\midrule
  \model &             + entropy loss   &        512 &  50.4 &  44.0 &  64.5 &   57.8 \\
  \model &   + adv. augs                &        512 &  55.5 &  49.7 &  71.0 &   65.8 \\
  \model &                    + mixup   &        512 &  56.8 &  51.1 &  72.2 &   67.1 \\
\bottomrule
\end{tabular}
}
    \caption{
    Ablation from SimCLR to our method, showing DISC2021 \uAP performance  for models trained on ImageNet.
    To compare descriptors of different sizes, we also show metrics after PCA reduction to 256 dimensions.
    \label{tab:imagenet_ablations}
    }

\end{table}

\paragraph{Entropy weight.}

Table \ref{tab:ssnpp_entropy_ablation} compares how varying entropy
loss weight ($\lambda$) affects copy detection accuracy, using
\SimCLRCD as a baseline. Models for this experiment are trained using
the strong blur augmentation setting.

\begin{table}[h]
    \centering
\scalebox{0.9}{
\begin{tabular}{lrr|rr}
\toprule
     model &    \uAP &  \uAPSN &  recall@1 &    MRR \\
\midrule
\SimCLRCD      &  33.0 &   51.6 &     58.6 &  60.5 \\ %
$\lambda = 1$  &  33.1 &   51.9 &     58.7 &  60.9 \\ %
$\lambda = 3$  &  38.0 &   56.1 &     62.9 &  65.1 \\ %
$\lambda = 10$ &  45.3 &   61.5 &     67.7 &  69.5 \\ %
$\lambda = 30$ &  50.4 &   64.5 &     69.8 &  71.4 \\ %
\bottomrule
\end{tabular}
}
\caption{
    DISC2021 accuracy metrics with varying entropy weight $\lambda$
    for models trained on ImageNet.
    \label{tab:ssnpp_entropy_ablation}
}
\end{table}

As the entropy weight increases, we see a corresponding increase in
global accuracy metrics. We also see a similar increase in per-query
ranking metrics, such as recall at 1 and mean reciprocal rank (MRR).
The increase in ranking metrics demonstrates that differential
entropy regularization improves copy detection accuracy in general,
beyond creating a more uniform notion of distance.

In contrast to metric learning contexts where entropy regularization
has been used, 
copy detection benefits from higher $\lambda$ values. %
Our standard setting is $\lambda = 30$, while \cite{elnouby2021vitretrieval}
reports reduced accuracy with $\lambda > 1$, and \cite{sablayrolles2019spreading}
uses values $< 0.1$. 
At $\lambda>40$, training becomes unstable, and tends to minimize the entropy loss at the
expense of the InfoNCE loss: embeddings are uniformly
distributed, but meaningless because image copies are not near anymore. 

\paragraph{Additional ablations.}

We explore how batch size, training schedule, descriptor dimensions, and score normalization affect accuracy in Appendix \ref{sec:additional_ablations}.

\section{Discussion}

\label{sec:discussion}
\paragraph{Dimensional collapse.}

We find, similar to \cite{zbontar2021barlow,jing2021collapse},
that SimCLR %
collapses to a subspace of approximately 256 dimensions when trained in 512 dimensions.
Table~\ref{tab:imagenet_ablations} shows that SimCLR's accuracy does
not improve much when the descriptor size increases from 128 to 512 dimensions. 
\model's entropy regularization resolves this collapse,
and allows the model to use the full descriptor space.

\paragraph{Entropy regularization and whitening.}

\model is much more accurate than baselines when compared
\emph{without} whitening or similarity normalization: 47.8 \uAP for $\lambda = 30$
when trained on ImageNet, versus 26.8 for an equivalent $\lambda = 0$ model.
Both the entropy loss and post-training PCA whitening aim at creating a more
uniform descriptor distribution. 
However PCA whitening can distort the descriptor space learned during training, particularly when many dimensions have trivial
variance.
Differential entropy regularization promotes an approximately uniform space, allowing the model to adapt to an approximately whitened descriptor during training, reducing the distortion whitening induces.

\paragraph{Uniform distribution as a perceptual prior.}

For most experiments in this work we focus on the \uAP metric that requires a separation between matches and non-matches at a fixed threshold.
However Table~\ref{tab:ssnpp_entropy_ablation} shows that ranking metrics \emph{also} improve with increased the entropy loss weight, \ie better calibration across queries does not fully explain the benefit of entropy regularization.

Differential entropy regularization acts as a kind of prior, selecting for an embedding space that is uniformly distributed.
We argue that, when applied to contrastive learning, this regularization is a \textbf{perceptual prior}, selecting for stronger copy detection representations.
An ideal copy detection descriptor would map copies of the same image together, while keeping even semantically similar (same ``class'') images far apart %
\ie the descriptor distribution is uniform.
This differs from the ideal properties of a representation for transfer learning to classification, where images depicting the same class should be nearby (a dense region) and well separated other classes (a sparse region between classes).

\paragraph{Visual results.}

Figure~\ref{fig:imageexamples} shows a few retrieval results, where \model outperforms the vanilla SimCLR. 
The two first examples demonstrate the impact of more appropriate data augmentation at training time: \model ignores text overlays and blur/color balance. 
The two last examples show that SimCLR falls back on low-level texture matching (grass) when \model correctly recovers the source image.

\paragraph{Limitations.}

Our method is explicitly text-insensitive when training with text augmentation, and we find that it is somewhat text-insensitive even when trained without text augmentation.
For this reason, \model is not precise when matching images composed entirely of text.
Different photos of the same scene (\eg of landmarks) may be identified as copies, even if the photos are distinct.
Sometimes, images are combined to create a composite image or collage, where the copied content may occupy only a small region of the composite image.
``Partial'' copies of this kind are hard to detect with global descriptor models like \modelx, and local descriptor methods may be necessary in this case.
Finally, matching at high precision often requires an additional verification step.

\paragraph{Ethical considerations.}

We focus our investigation on the DISC2021 dataset, which is thoughtful in its approach to images of people, using only identifiable photos of paid actors who gave consent for their images to be used for research.
Copy detection for content moderation is adversarial.
There is a risk that publishing research for this problem will better inform actors aiming to evade detection.
We believe that this is offset by the improvements that open research will bring.

This technology allows scaling manual moderation, which helps protect users form harmful content.
However, it can also be used for \eg political censorship. 
We still believe that advancing this technology is a net benefit.

\section{Conclusion}

We presented a method to train effective image copy detection models.
We have demonstrated architecture and objective changes to adapt contrastive learning to copy detection.
We show that 
the differential entropy regularization dramatically improves copy detection accuracy, promoting consistent separation of image descriptors.

Our method demonstrates strong results on DISC2021, significantly surpassing baselines, and transfers to Copydays, yielding state-of-the-art results.
Our method is efficient because it relies on a standard trunk, uses smaller inference sizes than are typical for image retrieval, and produces a compact descriptor.
Additionally, its calibrated distance metric limits candidates for verification.
We believe that these results demonstrate a unique compatibility between uniform embedding distributions and the task of copy detection.

{\small
\bibliographystyle{ieee_fullname}
\bibliography{biblio}
}

\clearpage \newpage

\appendix\newpage

\ifnotarxiv

\begin{center}
{\LARGE
Supplementary material for \\[0.2cm]
\textbf{A Self-Supervised Descriptor}
\textbf{for Image Copy Detection}

\vspace{0.5cm} %
}
\end{center}

\fi

\ifarxiv 

\begin{center}
{\LARGE Appendix }
\end{center}

\fi

We provide more details about the ablations (Appendix~\ref{sec:additional_ablations}) and Copydays results (Appendix~\ref{sec:copydays_full}). 
We also report a few additional details about the embedding distribution (Appendix~\ref{sec:embeddingdis}) and implementation details (Appendix~\ref{sec:implemdetails}). 
The last appendix~\ref{sec:imageres} shows additional example matches. 

\begin{table*}[b]
\centering
\begin{tabular}{r|rr}
\toprule
 batch size &    \uAP &  \uAPSN \\
\midrule
       2048 &  54.4 &   67.7 \\
       4096 &  56.6 &   69.2 \\
       8192 &  58.2 &  70.0 \\
      16384 &  {\bf 59.4} & {\bf 70.2} \\
\bottomrule
\multicolumn{3}{c}{~} \\
\end{tabular}
~~~
\begin{tabular}{r|rr}
\toprule
 epochs &   \uAP &  \uAPSN \\
\midrule
           25 &  54.4 &    67.4 \\
           50 &  56.2 &    68.9 \\
          100 &  {\bf 56.6} &    {\bf 69.2} \\
          200 &  56.3 &    68.9 \\
          400 &  55.7 &    68.1 \\
\bottomrule
\end{tabular}
~~~
\begin{tabular}{r|r|rr}
\toprule
 dimensions &    \uAP &  \uAPSN &  \uAPSN 256d \\
\midrule
        128 &  49.4 &   59.4 &        59.4 \\
        256 &  53.9 &   65.6 &        {\bf 65.6} \\
        512 &  56.6 &   69.2 &        64.0 \\
       1024 &  {\bf 57.3} &   {\bf 70.9} &        62.8 \\
       2048 &  56.8 &   70.8 &        62.9 \\
\bottomrule
\end{tabular}

\caption{
    Impact of three training parameters on the accuracy: 
    batch size, number of epochs and dimensionality. 
    We report \uAP performance on DISC21 for \model including advanced augmentations and $\lambda=15$, with and without score normalization. 
    For the dimensionality experiment we additionally report the accuracy after reduction to 256 dimensions.
    \label{tab:various_ablations}
}
\end{table*}

\section{Additional ablations}

\label{sec:additional_ablations}

Table~\ref{tab:various_ablations} shows how copy detection accuracy
is affected by several hyper-parameters.

\paragraph{Descriptor dimensionality.}

The descriptor dimension is a tradeoff between accuracy and the efficiency of the retrieval step.
When constraining the descriptor to 256 dimensions for retrieval, we see highest accuracy for descriptors trained at that size.

\paragraph{Batch size.}

The training objective learns to match pairs within the global batch (across all GPUs).
A larger batch size makes the training task more challenging, improving the final accuracy.
Large batch sizes require training with more machines, and incur synchronization overhead due in part to synchronized batch normalization.

\paragraph{Training schedule.}

We compare accuracy as we vary the number of training epochs, and find no benefit to longer training schedules.

\paragraph{Variance between initializations.}

We train using the same setting, initializing the model with five random seeds, and find a standard deviation of 0.2\% \uAP and 0.1\% \uAPSN.

\paragraph{Similarity normalization settings.}

We show score normalized accuracy given several similarity normalization settings
in Table~\ref{tab:score_normalization_ablation}.
Several score normalization settings work similarly well.
When using a single neighbor to normalize similarity, using the $2$nd nearest neighbor works best ($n = 2$).
When using an average similarity across multiple neighbors, averaging the first 2, 3 or 4 neighbors work similarly well.
We find that $\beta = 1$ is a good normalization weight.
Our similarity normalized results use $n = 1$, $n_{end} = 3$, $\beta = 1$, a setting that we found to work well across many descriptors.

\begin{table}[h]
\begin{tabular}{lr||lr||lr}
\toprule
\multicolumn{2}{c|}{$\beta=1, n=\nend$} & 
\multicolumn{2}{c|}{$\beta=1, n=1$} & 
\multicolumn{2}{c}{$n=1, \nend=3$} \\ 
\midrule
$n$ & \uAP &
$\nend$ & \uAP &
$\beta$ & \uAP \\
\midrule
 1 &  69.5 &      1 &  69.5 &  0.50 &  68.4 \\
 2 &  {\bf 71.1} &      2 &  71.0 &  0.75 &  70.4 \\
 3 &  70.8 &      3 &  {\bf 71.1} &  1.00 &  {\bf 71.1} \\
 4 &  70.3 &      4 &  {\bf 71.1} &  1.25 &  {\bf 71.1} \\
 5 &  69.7 &      5 &  71.0 &  1.50 &  70.6 \\
\bottomrule
\end{tabular}
\caption{
    DISC2021 \uAP with different score normalization settings for a \model trained on DISC2021 with advanced augmentations.
    \label{tab:score_normalization_ablation}
}
\end{table}

\paragraph{Trunk and projected features.}

We compare \model trunk and projected features in Table~\ref{tab:trunk_proj}.
Using the linear projection at inference time improves accuracy, despite a significantly more compact code.

\begin{table}[h]
\centering
\begin{tabular}{lrrr}
\toprule
descriptor & dims &   \uAP &  \uAPSN \\
\midrule
     trunk & 2048 &  57.2 &    71.9 \\
 projected &  512 &  {\bf 61.5} &    {\bf 72.5} \\
\bottomrule
\end{tabular}
\caption{
    DISC2021 accuracy of \model trunk and projected trained on DISC2021 with advanced + mixup augmentations.
    \label{tab:trunk_proj}
}
\end{table}

\section{Full Copydays results}

We provide additional Copydays results in Table~\ref{tab:copydays_full}, evaluating \model and \modellarge using preprocessing settings from prior published results.
In each case, we evaluate our method with no tuning, \eg we don't adjust the GeM $p$ as proposed in \cite{berman2019multigrain}.

\label{sec:copydays_full}

\begin{table}[h]
\centering
\resizebox{\columnwidth}{!}{
\begin{tabular}{llrlrr}
\toprule
      model &     trunk & dims &   size &  $mAP$  & \uAP \\
\midrule
 Multigrain~\cite{berman2019multigrain} &  ResNet50 &       1500 &  long 800 &  82.3 &  77.3 \\
       DINO~\cite{caron2021emerging} &  ViT-B/16 &       1536 &           $224^2$ &  82.8 &  92.3 \\
       DINO~\cite{caron2021emerging} &   ViT-B/8 &       1536 &           $320^2$ &  86.1 &  88.4 \\
\midrule
       \model &  ResNet50 &        512 &           $224^2$ &  84.9 &  98.3 \\
       \model &  ResNet50 &        512 &           $320^2$ &  87.4 &  98.3 \\
       \model &  ResNet50 &        512 &  short 288 &  86.6 &  98.1 \\
       \model &  ResNet50 &        512 &  long 800 &  90.0 &  93.9 \\
\midrule
       \modellarge &  ResNeXt101 &     1024 &  $224^2$ &  87.3 & 98.6 \\
       \modellarge &  ResNeXt101 &     1024 &  $320^2$ &  90.6 &  98.6 \\
       \modellarge &  ResNeXt101 &     1024 &  short 288 & 91.8 & \bf{98.7} \\
       \modellarge &  ResNeXt101 &     1024 &  long 800 &  \bf{93.6} &  97.1 \\
\bottomrule
\end{tabular}
} %

    \caption{
    Full Copydays (CD10K) results: accuracy measured in mAP on the ``strong'' subset,
    and \uAP on the full dataset.
    \label{tab:copydays_full}
    }
\end{table}

We note that at $224^2$ inference size, ResNet50 has approximately $4 \times$ the throughput as ResNeXt101 or ViT-B/16, and $20\times$ that of ViT-B/8.\cite{caron2021emerging}

\section{Embedding distribution}
\label{sec:embeddingdis}

\begin{figure}
    \centering
    \includegraphics[width=\columnwidth]{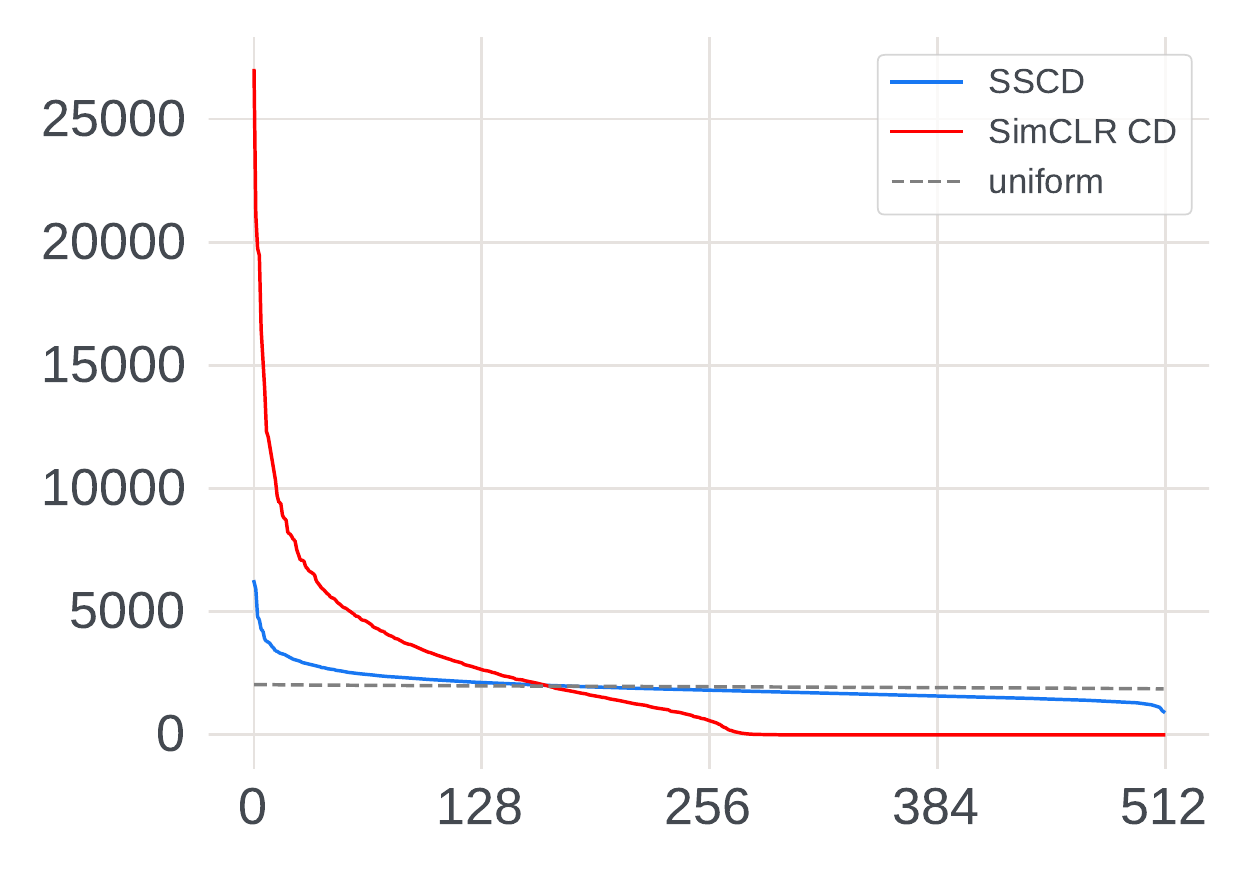}  %
    \caption{
    Descriptor principal values on the DISC2021 reference set: {\color{blue}\model} ($\lambda = 30$) and {\color{red}SimCLR$_{CD}$} ($\lambda = 0$), compared to a reference uniform distribution. %
    }
    \label{fig:principal_values}
\end{figure}

We plot principal values for \model ($\lambda = 30$) compared to SimCLR$_{CD}$ ($\lambda = 0$), and a uniform distribution in Figure~\ref{fig:principal_values}.
We see that the $\lambda = 0$ model fails to make full use of the descriptor space, as observed in \cite{zbontar2021barlow,jing2021collapse}.
With entropy regularization, all components have similar energy, spanning less than an order of magnitude (the maximum is $6.6\times$ the minimum).

\section{Implementation details}
\label{sec:implemdetails}

\label{sec:implementation_details}

\paragraph{Mixup and Cutmix.}

Mixup and Cutmix augmentations both combine content from two source images.
The amount of content used from each image is determined by a mixing parameter $\gamma$, sampled from a $\beta$ distribution: $\gamma \sim \beta(\alpha, \alpha)$.
We set $\alpha = 2$ to reduce the prevalence of ``trivial'' mixed images that draw nearly all content from one of the inputs.

\paragraph{DINO baseline details.}

We follow the copy detection method presented in \cite{caron2021emerging} for the DINO baseline.
We use the concatenation of the CLS token and GeM pooled ($p=4$) patch token features as the descriptor.

Our DINO DISC evaluation uses the ViT-B/16 trunk.
We resize inputs to $224 \times 224$ without center cropping.
This outperformed other preprocessing for this model, including our default aspect-ratio preserving resize, and resizing inputs to a larger fixed size ($288 \times 288$).
We suspect that ViT models may be less adaptable to rectangular inputs than fully convolutional networks.

\section{Visualizing matches}
\label{sec:visumatches}

To view which parts of an image A match strongly to another image B, we experiment by keeping the activation map on A at full resolution by removing the GeM pooling operation. 
This results into one descriptor per activation map pixel, that can be compared with a global \model descriptor. 
We can thus build a spatial heatmap with the strongest activations. 
Figure~\ref{fig:heatmaps} shows image pairs and the corresponding heatmaps.
The areas on the left image that match with the image on the right are clearly identified.

\newcommand{\igheatmap}[3]{
    \includegraphics[width=0.3\columnwidth,align=c]{fig/heatmaps/#1_#2/#1.jpg}~~%
    \includegraphics[width=0.3\columnwidth,align=c]{fig/heatmaps/#1_#2/heatmap16.png}~~%
    \includegraphics[width=0.3\columnwidth,align=c]{fig/heatmaps/#1_#2/#2.jpg}%
    \imcreds{#3}
}

\begin{figure}
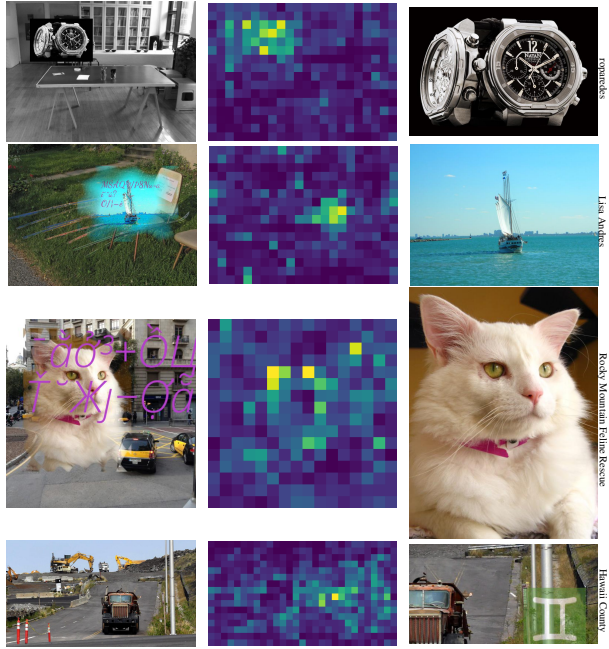

\centering
\igheatmap{Q00037}{R450692}{roparedes} \\
\igheatmap{Q00550}{R961078}{Lisa Andres} \\
\igheatmap{Q00846}{R033260}{Rocky Mountain Feline Rescue} \\
\igheatmap{R075820}{Q00667}{Hawaii County} \\ 
    \caption{Left and right columns: Pairs of matching images from the DISC2021 dataset. 
    The central column shows which areas of the left image match best with the image on the right:
    yellow is strong match, blue is neutral or negative.}
    \label{fig:heatmaps}
\end{figure}

\section{Retrieved matches}
\label{sec:imageres}

We compare the first result retrieved by \model and SimCLR on the DISC2021 dataset.
Both models are trained on ImageNet and evaluated with whitening.
We use trunk features for SimCLR, which are more accurate for this model.
We do not use score normalization, since it has no effect on top-1 accuracy.

\newcommand{\cmark}{\ding{51}}%
\newcommand{\xmark}{\ding{55}}%

\begin{table}[h]
\centering
\begin{tabular}{llr}
\toprule
 \model &  SimCLR & queries \\
\midrule
        \cmark &           \cmark &  38.9 \% \\
        \cmark & \xmark &  39.0 \% \\
        \xmark &           \cmark &    0.3 \% \\
        \xmark &           \xmark &  21.8 \% \\
\bottomrule
\end{tabular}
    \caption{
    Percentage of DISC2021 query first result accuracy by model for \model and SimCLR trained on ImageNet.
    \label{tab:recall_at_one_counts}
    }
\end{table}

Table~\ref{tab:recall_at_one_counts} shows quantitative results from this exercise.
\model correctly identifies the copy as the first result $2\times$ as often as SimCLR.
Correct \model matches are nearly a superset of SimCLR matches: very rarely does SimCLR have a correct first result that \model misses.

Figure~\ref{fig:more_image_examples} shows additional queries and retrieved results for examples that only \model correctly identifies.
One pattern we observe is that SimCLR often matches images with similar types of distortion together.
Images with text at an angle, or strong diagonal features, may be incorrectly matched with images with similar features.
Images with a blurry, or grainy, quality are matched to other images with a similar quality.
This is surprising given that SimCLR trains with a blur augmentation, albeit weaker, and should be somewhat blur invariant.

\newcommand{\morecomp}[1]{\includegraphics[width=0.34\columnwidth,align=c]{fig/more_retrieval_examples/#1.jpg}}
\newcommand{\morecompv}[1]{\includegraphics[height=0.34\columnwidth,align=c]{fig/more_retrieval_examples/#1.jpg}}

\begin{figure}
    \centering
    \hspace*{-1em}
\begin{tabular}{c@{\hspace{1mm}}c@{\hspace{1mm}}c}
    \toprule
    Query  & \model & SimCLR \\
    \midrule

\smallskip   \morecomp{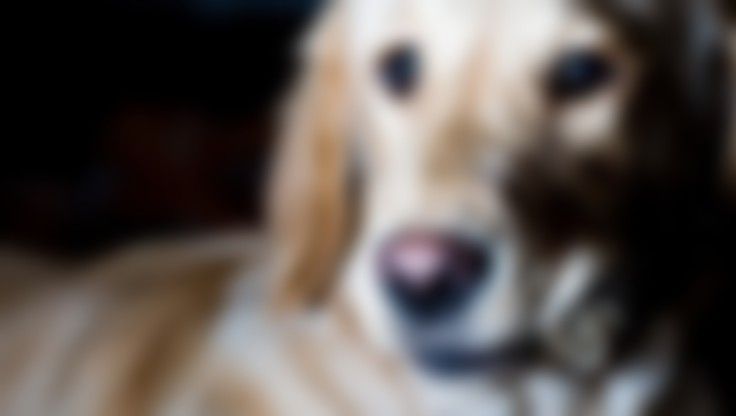} &   \morecomp{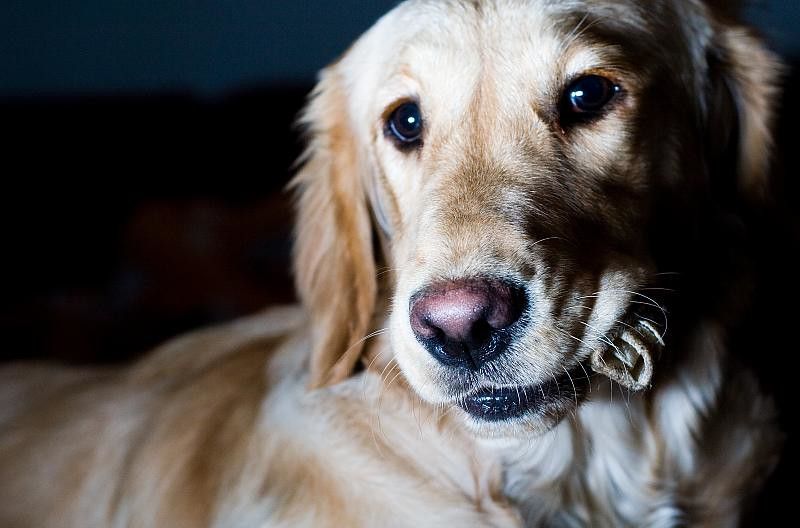}\imcreds{aranyember} &   \morecomp{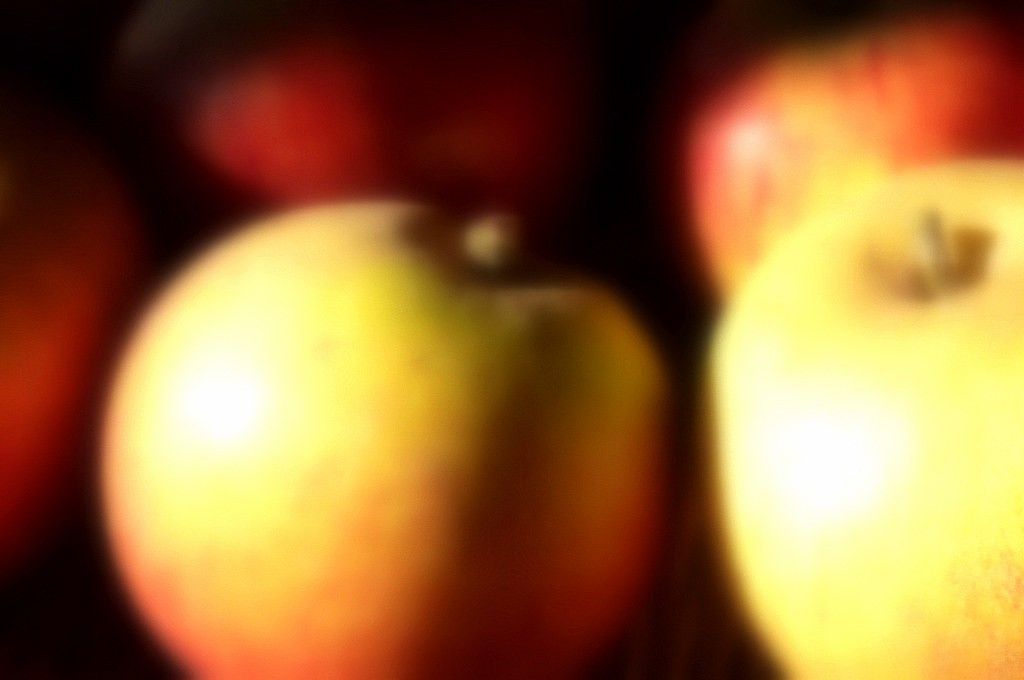}\imcreds{muffinn} \\
\smallskip   \morecomp{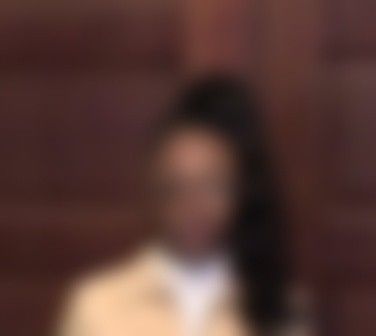} &   \morecomp{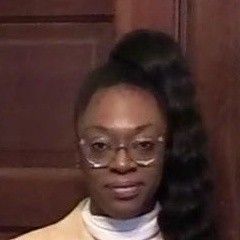}\imcreds{DFDC} &   \morecomp{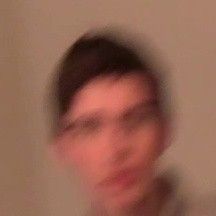}\imcreds{DFDC} \\
\smallskip  \morecompv{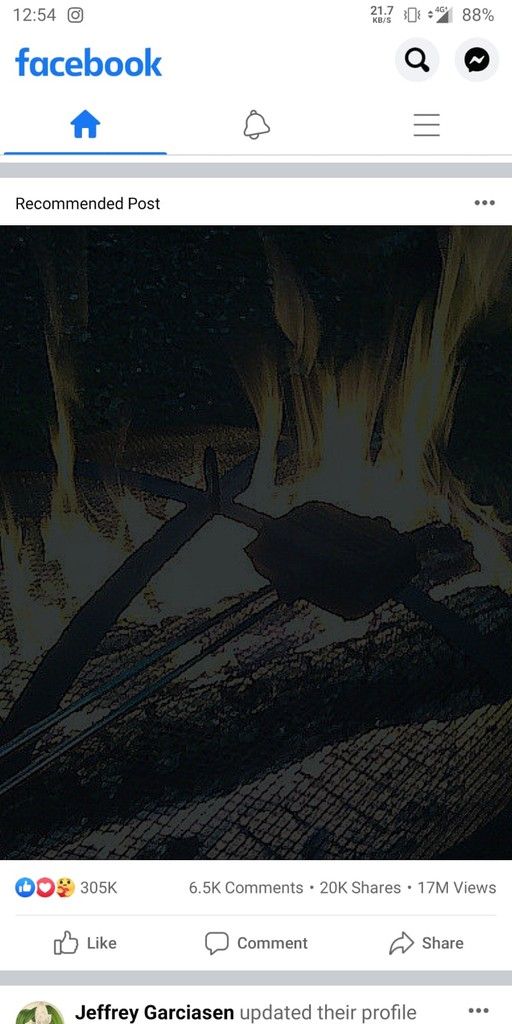} &   \morecomp{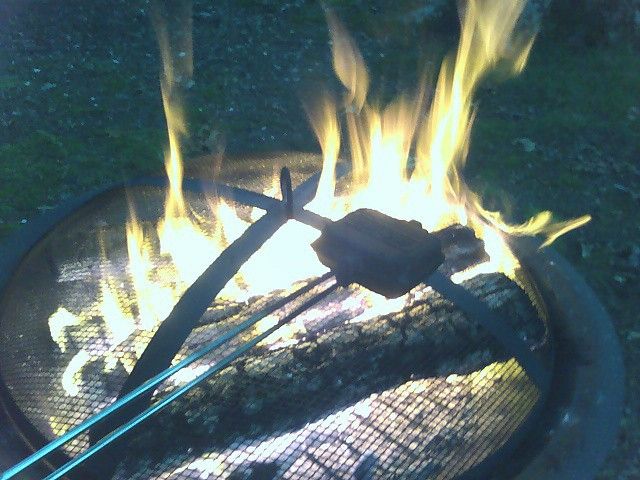}\imcreds{kjarrett} &  \morecompv{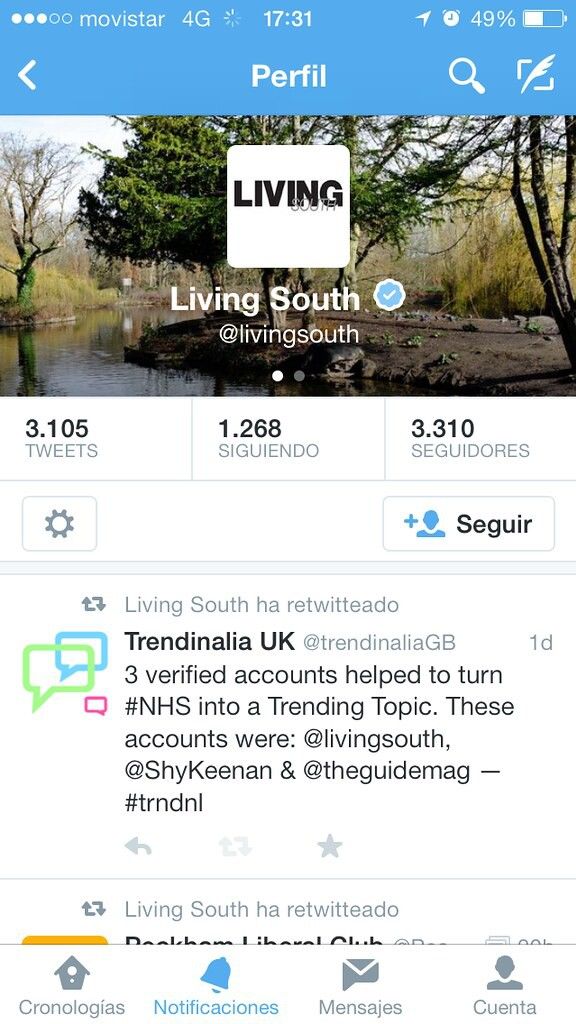}\imcreds{La Conversa} \\
\smallskip  \morecompv{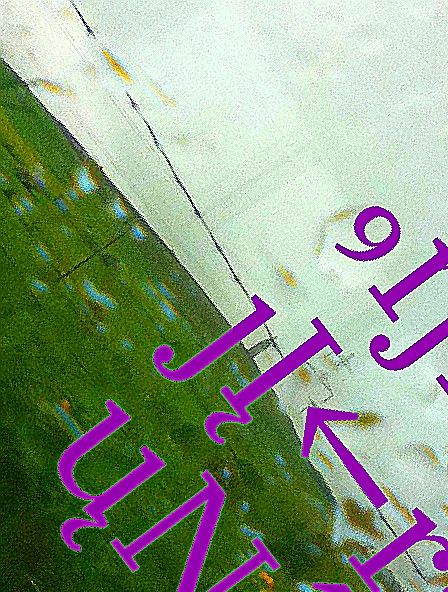} &   \morecomp{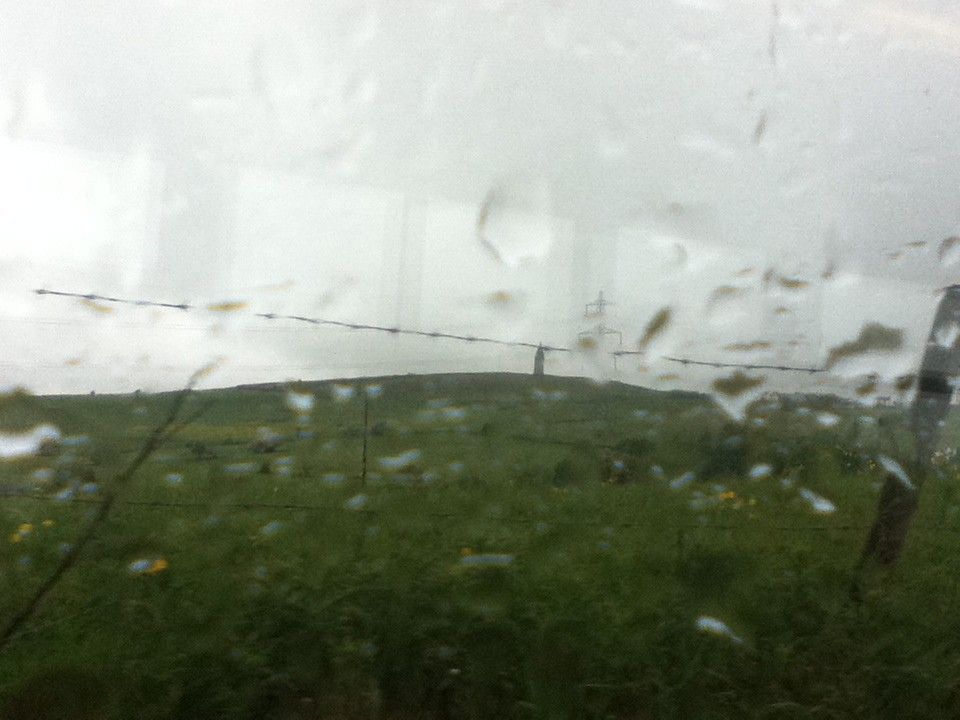}\imcreds{Gene Hunt} &   \morecomp{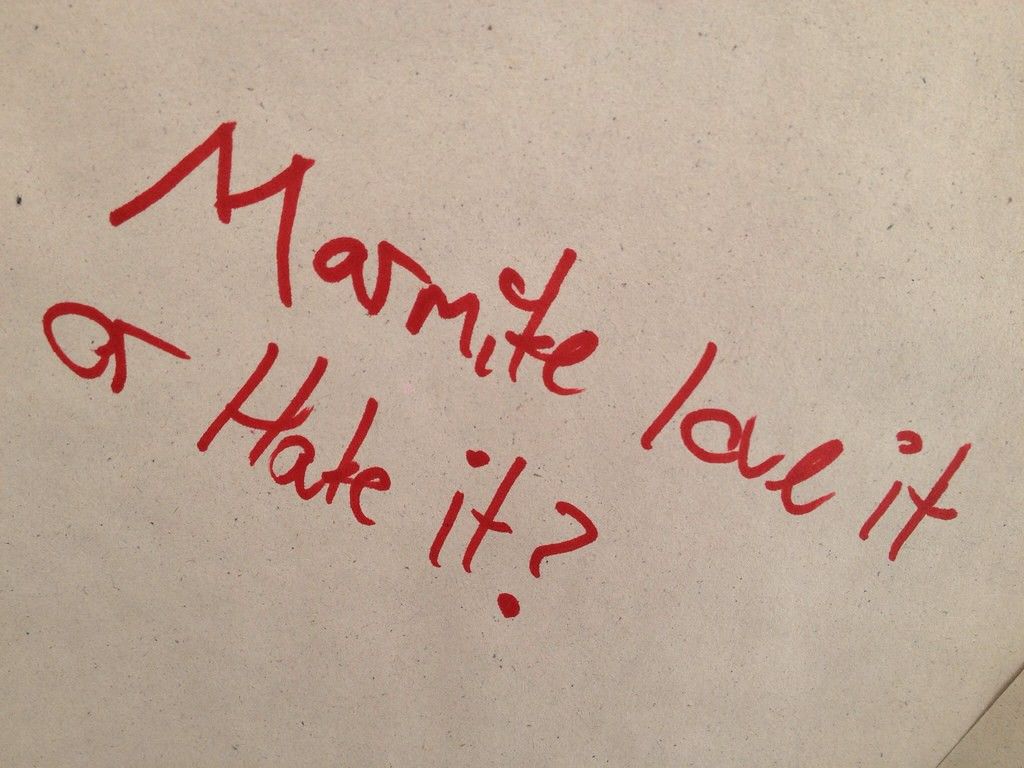}\imcreds{markheybo} \\
\smallskip   \morecomp{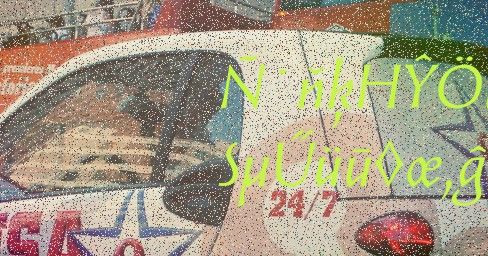} &   \morecomp{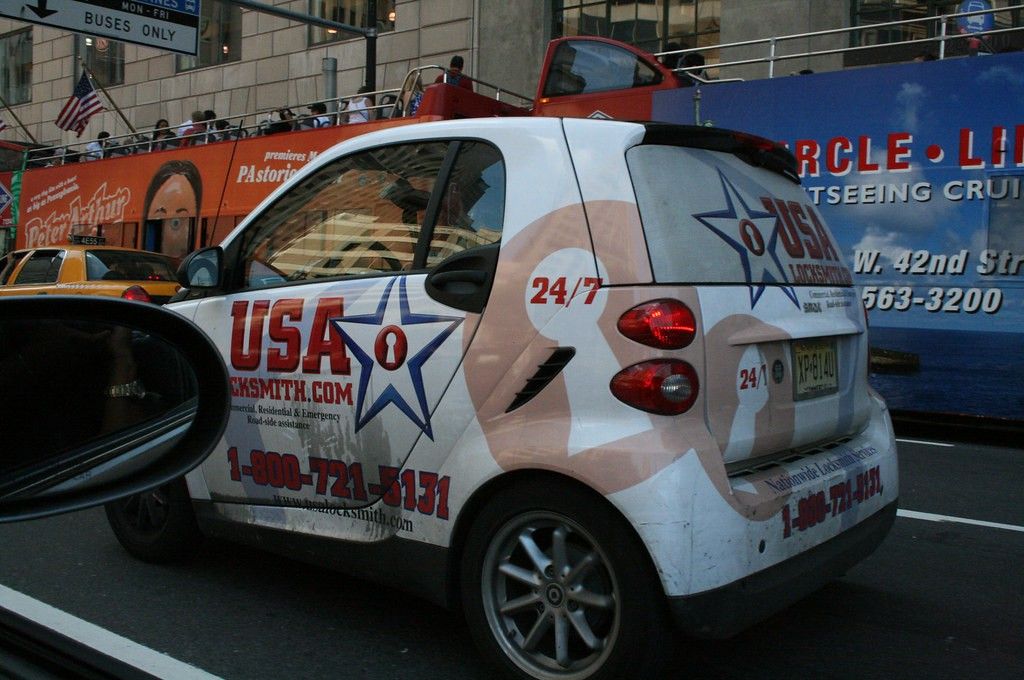}\imcreds{jturbanos} &   \morecomp{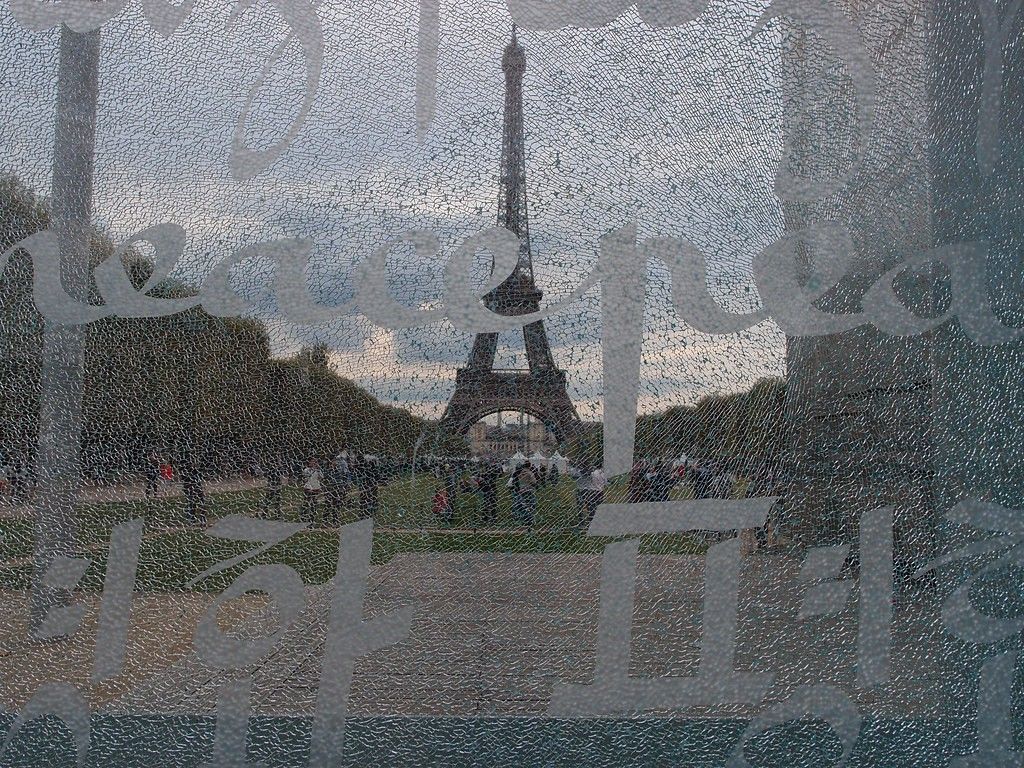}\imcreds{*\_*} \\
\smallskip   \morecomp{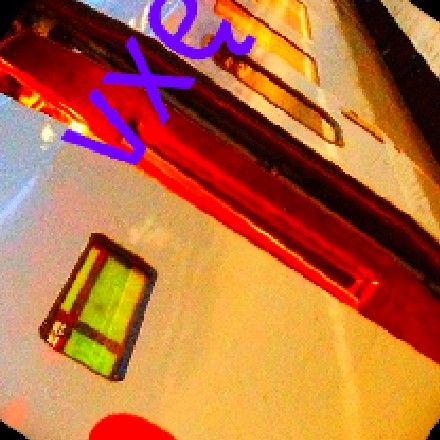} &   \morecomp{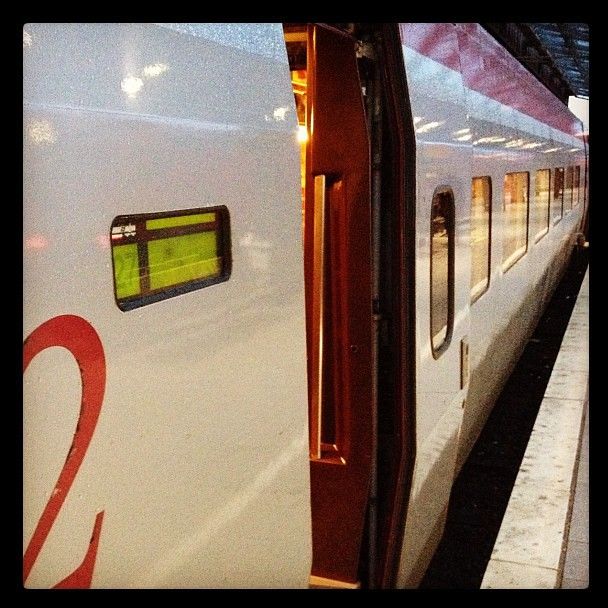}\imcreds{hirnrinde} &   \morecomp{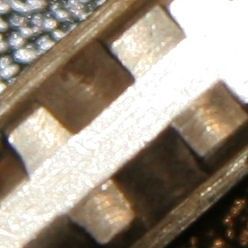}\imcreds{exfordy} \\
\smallskip  \morecompv{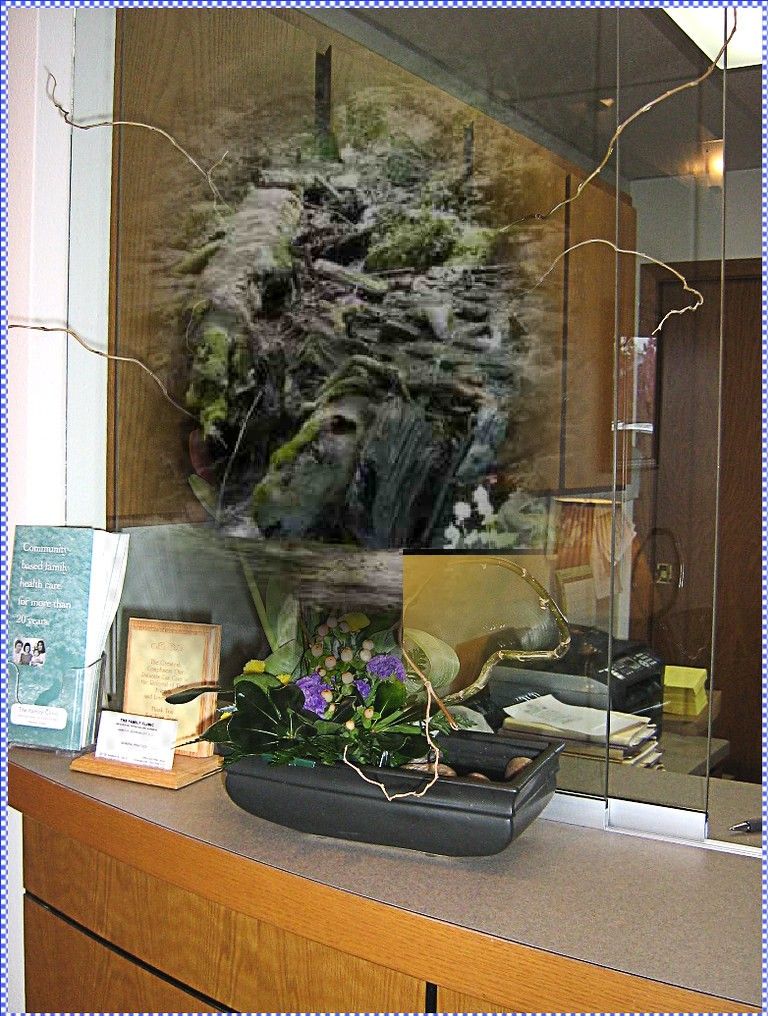} &  \morecompv{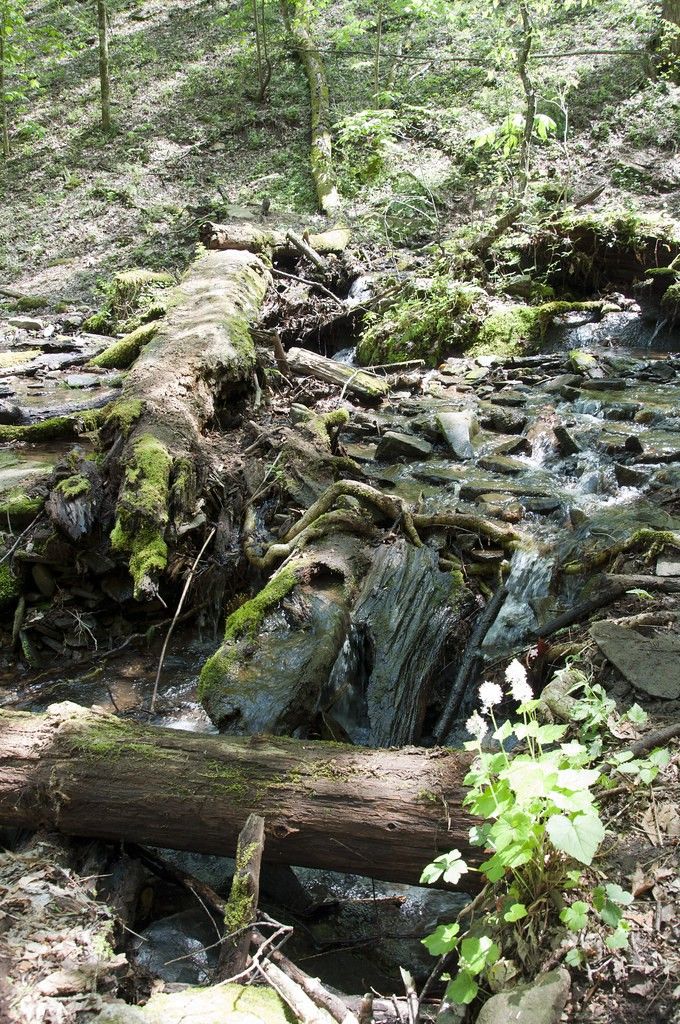}\imcreds{John Philip Henry} &   \morecomp{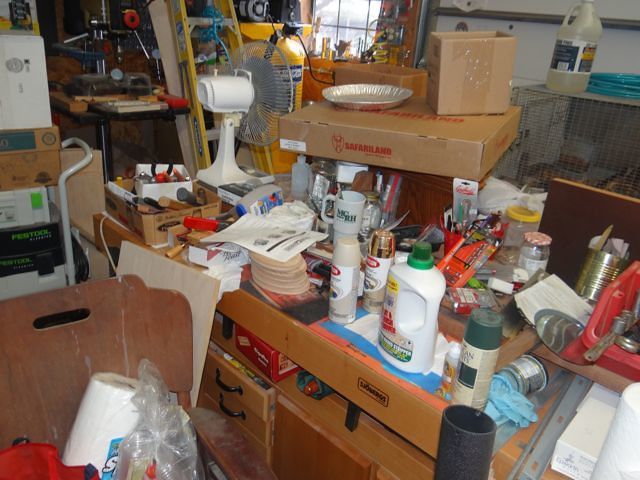}\imcreds{DakilAuctioneers} \\
    \bottomrule
\end{tabular}
    \caption{
    Example retrieval results from the DISC2021 dataset.
    For each row, we show the query image, the top retrieval result for \model, the top retrieval result for SimCLR. 
    }
    \label{fig:more_image_examples}
\end{figure}

\end{document}